\documentclass[12pt]{article}

\usepackage[allcolors=blue,colorlinks=true]{hyperref}

\usepackage[margin=1in]{geometry}

\usepackage{float}

\makeatletter
\renewcommand*{\@fnsymbol}[1]{\@alph{#1}}
\makeatother

\usepackage{natbib}
\bibpunct{(}{)}{;}{a}{}{,}

\usepackage{microtype}
\usepackage{graphicx}
\usepackage{subfigure}
\usepackage{booktabs} 

\usepackage{xcolor}

\usepackage{amsmath,amssymb}
\DeclareMathOperator{\bTheta}{\Theta}

\DeclareMathOperator*{\argmin}{argmin}
\DeclareMathOperator*{\tr}{tr}

\usepackage{amsfonts}

\usepackage{hyperref}

\usepackage{setspace}  

\usepackage{thmtools}
\usepackage{thm-restate}
\usepackage{cleveref}

\declaretheorem{assumption}
\declaretheorem{theorem}
\declaretheorem{lemma}
\declaretheorem{proposition}

\usepackage{accents}
\newcommand{\ubar}[1]{\underline{#1\mkern-4mu}\mkern4mu }

\newcommand{\hide}[1]{}
\newcommand{\qed}{\hfill \blacksquare}

\newcommand{\vertiii}[1]{{\left\vert\kern-0.25ex\left\vert\kern-0.25ex\left\vert #1 
    \right\vert\kern-0.25ex\right\vert\kern-0.25ex\right\vert}}

\begin{document}


\title{\bf Thresholded Graphical Lasso Adjusts for Latent Variables: Application to Functional Neural Connectivity}
\author{ Minjie Wang\thanks{Department of Statistics, Rice University, Houston, TX},\hspace{.2cm}
and Genevera I. Allen\thanks{Departments of Electrical and Computer Engineering, Statistics, and Computer Science, Rice University, Houston, TX} \textsuperscript{,}\thanks{Jan and Dan Duncan Neurological Research Institute, Baylor College of Medicine, Houston, TX}}
\date{}
\maketitle

\begin{abstract}
In neuroscience, researchers seek to uncover the connectivity of neurons from large-scale neural recordings or imaging; often people employ graphical model selection and estimation techniques for this purpose.  But, existing technologies can only record from a small subset of neurons leading to a challenging problem of graph selection in the presence of extensive latent variables.  \citet{chandrasekaran2012} proposed a convex program to address this problem that poses challenges from both a computational and statistical perspective.  To solve this problem, we propose an incredibly simple solution: apply a hard thresholding operator to existing graph selection methods.  Conceptually simple and computationally attractive, we demonstrate that thresholding the graphical Lasso, neighborhood selection, or CLIME estimators have superior theoretical properties in terms of graph selection consistency as well as stronger empirical results than existing approaches for the latent variable graphical model problem.  We also demonstrate the applicability of our approach through a neuroscience case study on calcium-imaging data to estimate functional neural connections.  
\end{abstract}

\noindent%

\newpage

\doublespacing

\section{Introduction}
\label{intro}

Emerging neuroscience technologies such as electrophysiology and calcium imaging can record from tens-of-thousands of neurons in the live animal brain while the animal is responding to stimuli and behaving freely.  Scientists often seek to understand how neurons are communicating during certain stimuli or activities, something termed functional neural connectivity.  To learn functional connections from large-scale neuroscience data, many have proposed using probabilistic graphical models \citep{yatsenko2015improved,narayan2015two,chang2019graphical}, where each edge denotes conditional dependencies between nodes.  Yet, applying such models in neuroscience poses a major challenge as only a small subset of neurons in the animal brain can be recorded at once, leading to abundant latent variables.  \citet{chandrasekaran2012} termed this the latent variable graphical model problem and proposed a convex program to solve this.  While conceptually attractive, this approach poses several statistical, computational and practical challenges, discussed subsequently, for the task of learning functional neural connectivity from large-scale neuroscience data. Because of this, we are motivated to consider an incredibly simple solution to the latent variable graphical model problem: apply a hard thresholding operator to existing graph selection estimators.  In this paper, we study this approach showing that thresholding has more desirable theoretical properties as well as superior empirical performance.

\subsection{Related Work}



Let $X=\left(X_1, \ldots, X_p\right)$ be a $p$-dimensional random vector following a multivariate normal distribution $N(\mu,\Sigma)$ with mean vector $\mu$ and covariance matrix $\Sigma$. Denote the precision matrix $\bTheta = \Sigma^{-1}$. 
Gaussian graphical model, represented by $G = (V,E)$, has been widely used to uncover dependence structure, in which the vertices or nodes $V$ index a collection of random variables and edges $E$ represent conditional independence between variables. An edge in $E$ connects two vertices $(i,j)$ if and only if $\bTheta_{ij} \neq 0$. The absence of an edge between $X_i$ and $X_j$ suggests that $X_i$ and $X_j$ are independent conditional on the other variables. In this way,  identifying the conditional dependence is equivalent to finding the non-zero entries in the precision matrix. 

In literature, many have studied model selection in the context of Gaussian graphical model. \citet{yuan2007model,banerjee2008model,friedman2008sparse} proposed to use the regularized maximum log-likelihood and apply $\ell_1$ penalty  to the off-diagonal entries of the precision matrix:
$$
\widehat{\bTheta}^{\lambda} =\argmin_{\bTheta \succ 0}\left\{\tr(\widehat{\Sigma} \Theta)-\log \det (\Theta)+\lambda \sum_{j \neq k}\left|\Theta_{j k}\right|\right\} .
$$
Going beyond, a plethora of work have studied the convergence of graphical Lasso solution. \citet{ravikumar2011high} derived the convergence rate in elementwise $\ell_{\infty}$-norm of graphical Lasso estimate and hence proved model selection consistency. In particular, they require the $\alpha$-incoherence condition to guarantee that the edge set of the graphical Lasso solution does not include any false edges. \citet{rothman2008sparse} established convergence in Frobenius norm, with milder assumptions on the inverse covariance, namely a bound on the eigenvalues.

In addition, many have proposed alternatives to estimate the sparse precision matrix. \citet{cai2011constrained} considered a method of constrained $\ell_1$-minimization for inverse matrix estimation  (CLIME) and demonstrated strong statistical guarantees.
\citet{meinshausen2006high} proposed neighborhood selection based approach by performing Lasso of each node on its neighbors and using the support of the estimated regression coefficients to predict the edge set.




In terms of studying latent variables in the context of graphical models, \citet{chandrasekaran2012} studied the case where the observed and hidden variables are jointly Gaussian. Specifically, they partitioned the vertices into subsets $O$ and $H$ of observed and hidden variables. Then the inverse covariance matrix $\bTheta = \Sigma^{-1}$ can be written as
$\bTheta=\left(\begin{array}{cc}
\bTheta_{O} & \bTheta_{O H} \\
\bTheta_{H O} & \bTheta_{H}
\end{array}\right).$
The marginal concentration matrix $(\Sigma^*_O)^{-1}$ associated with the observed variables $X_O$, is given by the Schur complement: 
$$
\left(\Sigma_{O}^{*}\right)^{-1}=\bTheta_{O}^{*}-\bTheta_{O H}^{*}\left(\bTheta_{H}^{*}\right)^{-1} \bTheta_{H O}^{*},
$$
which is a difference between the sparse term $\bTheta_{O}^{*}$ and the term $\bTheta_{O H}^{*}\left(\bTheta_{H}^{*}\right)^{-1} \bTheta_{H O}^{*}$ summarizing the effect of marginalization over the latent variables $X_H$. The matrix $\bTheta_{O H}^{*}\left(\bTheta_{H}^{*}\right)^{-1} \bTheta_{H O}^{*}$ is low-rank if the number of latent, unobserved variables is small.

The goal is to recover the sparse conditional matrix $\bTheta_O^{*}$ based on observed variables $X_O$. To solve this, \citet{chandrasekaran2012} proposed a regularized maximum-likelihood decomposition framework, which approximates the sample covariance matrix by decomposing the concentration matrix into a sparse and low-rank matrix:
\begin{align*}
(\widehat{S}, \widehat{L}) &= \argmin _{S-L \succ 0, L \succeq 0} \bigg\{\operatorname{tr}\left((S-L) \widehat{\Sigma}_{O}\right)-\log \det(S-L)   +\lambda\big(\gamma \sum_{j \neq k}\left|S_{j k}\right|+\operatorname{tr}(L)\big)\bigg\}.
\end{align*}
Moreover, they established the consistency of the estimator with  $n \gtrsim p$  
samples and irrepresentability conditions. Yet, as mentioned by \citet{wainwright2012discussion}, such condition is restrictive compared to the results without latent variables. For example, \citet{ravikumar2011high} proved that the graphical Lasso estimate has an element-wise $\ell_{\infty}$-norm error of order  $\sqrt{\frac{\log p}{n}}$ and operator norm error of order $\sqrt{\frac{d^2 \log p}{n}}$ where $d$ is the degree of the graph. \citet{chandrasekaran2012} suggested that $n \gtrsim p$ samples is inherent to the latent variable graphical model problem to ensure spectral norm consistency for the low-rank component.

Another line of research focuses on thresholding the estimators from a statistical model.  
In regression setting, \citet{lounici2008sup,meinshausen2009lasso}  proved the sign consistency of thresholded Lasso estimator with a proper choice of the threshold; \citet{zhou2010thresholded,van2011adaptive} proposed multi-step thresholding procedure based on the Lasso and showed model selection consistency under the restricted eigenvalue condition; \citet{giurcanu2016thresholding} proposed thresholding least-squares solution. 
For covariance estimation, \citet{bickel2008covariance,rothman2009generalized,cai2011adaptive} proposed thresholding the sample covariance matrix and obtained rate of convergence. Some have shown the equivalence between simply thresholding the sample covariance matrix and the graphical Lasso estimator \citep{mazumder2012exact,sojoudi2016equivalence,fattahi2019graphical}. 

\subsection{Contribution}

We solve the latent variable graphical model problem by applying a hard thresholding operator to existing graph selection methods for the Gaussian graphical model.  
Our approach is inspired by \citet{vinci2019graph} who established a theoretical result showing that latent variables in Gaussian graphical model induce false positives.  Under certain probabilistic and graph theoretic conditions, the false positives are smaller in magnitude than the minimum edge weights of the graph; hence, they suggest eliminating them via thresholding which we explore in this paper.
\citet{ren2012} also proposed thresholding the CLIME estimator as a solution to the latent variable graphical model problem.  Despite this, however, no one has theoretically studied thresholding graphical model estimators, such as the graphical Lasso and neighborhood selection, let alone studied this for solving the latent variable problem.

In this paper, we theoretically study thresholded graphical model estimators both on their own and in the presence of latent variables.  We demonstrate that hard thresholding is not just conceptually and computationally attractive, but also requires less stringent assumptions to achieve graph selection consistency and has an improved sample complexity compared to the latent variable graphical model estimator.  Additionally, we conduct a thorough empirical study showing numerous advantages of thresholded graph selection estimators.  We conclude with a case study on estimating functional neural connectivity from calcium imaging data.

\section{Thresholded Graphical Lasso}

We first introduce the thresholded graphical Lasso estimator. Given a graphical Lasso estimator $\widehat{\bTheta}^{\lambda}$, the corresponding thresholded estimator is defined by
\begin{equation}
\widetilde{\bTheta}_{ij}^{\lambda,\tau} =\begin{cases}
\widehat{\bTheta}_{ij}^{\lambda}, & \text { if }\left|\widehat{\bTheta}_{ij}^{\lambda}\right|> \tau  , \\
0 & \text { elsewhere, }
\end{cases}
\end{equation}
where $\widehat{\bTheta}_{ij}^{\lambda}$ is the  graphical Lasso estimator with regularization parameter $\lambda$ while constant $\tau$ controls the level of threshold. We can easily apply this hard thresholding operator to other graph selection methods. We consider the thresholded graphical Lasso, CLIME and neighborhood selection estimators in this paper and demonstrate the theoretical properties in the following subsections. All proofs are given in the Supplementary Materials.

\subsection{Thresholded Graphical Lasso}
\label{tglasso}

Before we state our main result of graphical model selection consistency for the thresholded graphical Lasso, we make the following assumptions on our model. Denote $\varphi_{\max}(\cdot)$ and $\varphi_{\min}(\cdot)$ as the largest and smallest eigenvalues of a matrix. Denote $s$ as the total number of non-zero edges, i.e., $E\left(\Theta^{*}\right):=\left\{ (i, j) \in V \times V \mid i \neq j, \Theta_{i j}^{*} \neq 0\right\}$ and $s = |E \left(\Theta^{*}\right)|$. 

\begin{assumption}[restate = assumptionAone, name = ]
\label{assumptionA1}
$X_{i} \text { be i.i.d. } \mathcal{N}\left(\mathbf{0}, \Sigma^* \right)$.
\end{assumption}
\begin{assumption}[restate = assumptionAtwo, name = ]
\label{assumptionA2}
$\varphi_{\min }\left(\Sigma^*\right) \geq \underline{k}>0,$ or equivalently $\varphi_{\max }\left(\Theta^*\right) \leq 1 / \underline{k}$.
\end{assumption}
\begin{assumption}[restate = assumptionAthree, name = ]
\label{assumptionA3}
$\varphi_{\max }\left(\Sigma^*\right) \leq \bar{k}$.
\end{assumption}
\begin{assumption}[restate = assumptionAfour, name = ]
\label{assumptionA4}
Define the minimum signal strength:
\begin{equation*}
 \theta_{\min }:=\min _{(i, j) \in E\left(\Theta^{*}\right)}\left|\Theta_{i j}^{*}\right| > c_1 \sqrt{\frac{s \log p}{n}}.   
\end{equation*}
\end{assumption}

Assumptions~\ref{assumptionA1}-\ref{assumptionA3} are the same as the one stated in the result by \citet{rothman2008sparse}, which are required to establish convergence in Frobenius norm for graphical Lasso estimator. Assumption~\ref{assumptionA4} is the minimum signal strength condition usually stated for model selection consistency. 


\begin{lemma}[restate = lemmaone, name =\citealt{rothman2008sparse}]
\label{lemma1}
Let Assumptions~\ref{assumptionA1}-\ref{assumptionA3} be satisfied. If $\lambda \asymp \sqrt{\frac{\log p}{n}}$, there exists a  $c_2$ such that the graphical Lasso estimate $\widehat{\bTheta}^{\lambda}$ satisfies:
$$
\left\| {\widehat{\bTheta}}^{\lambda}-\Theta^{*}\right\|_{F} \leqslant c_{2}  \sqrt{\frac{(p+s) \log p}{n}} \;\;,  
$$
with probability at least $1 - b_1 \exp(-b_2 n \lambda^2)$ where $b_1$ and $b_2$ depend on $\bar{k}$ only.
\end{lemma}

\citet{ravikumar2011high} yielded the same Frobenius norm convergence rate $\sqrt{ (p+s) \log p /n}$, but better convergence rate in spectral norm with $\mathcal{O}\left(\sqrt{\frac{\min \left\{d^{2},(s+p)\right\} \log p}{n}}\right)$. Moreover, they established convergence in $\ell_{\infty}$-norm.  We will compare our results with theirs below. Note, as discussed by \citet{rothman2008sparse}, the worst part of the rate in Lemma~\ref{lemma1}, $\sqrt{p \log p /n}$, comes from estimating the diagonal. Since we are interested in edge recovery, it can be shown that we can get the rate of $\sqrt{s \log p /n}$ for off-diagonal parts. Hence, we have Proposition~\ref{proposition_rothman_off_diag}. (Or, we could use the correlation matrix rather than the covariance matrix.)

\begin{proposition}[restate = propositionrothmanoffdiag, name = ]
\label{proposition_rothman_off_diag}
Let Assumptions~\ref{assumptionA1}-\ref{assumptionA3} be satisfied. If $\lambda \asymp \sqrt{\frac{\log p}{n}}$, there exists a  $c_2$ such that the graphical Lasso estimate $\widehat{\Theta}^{\lambda}$ satisfies:
$$
\mathbb{P}\left(  \left\|\widehat{\Theta}_{\text{off}}^{\lambda}-\Theta^{*}_{\text{off}}\right\|_{F} \leq  c_{2}  \sqrt{\frac{s \log p}{n}}  \right) \geq 1  - b_1 \exp(-b_2 n \lambda^2)  \to 1.
$$
\hide{
$$
\mathbb{P}\left(  \left\|\widehat{\Theta}_{ij}^{\lambda}-\Theta^{*}_{ij} \right\|_{F} \leq  c_{2}  \sqrt{\frac{s \log p}{n}} \right) \geq 1  - b_1 \exp(-b_2 n \lambda^2)  \to 1,  \hspace{6mm} \forall i \neq j.
$$
}
\end{proposition}

\hide{
\begin{theorem}
\label{theorem1}
Let assumptions in Lemma~\ref{lemma1} hold true. Further, if Assumption~\ref{assumptionA4} holds true with $c_1> 2c_2$, the thresholded graphical Lasso estimate $\widetilde{\bTheta}^{\lambda,\tau}$ with threshold level $\tau = c_2 \sqrt{\frac{s \log p}{n}}$ satisfies:
\begin{align*}
    &\mathbb{P}\left(\operatorname{sign}(\widetilde{\bTheta}_{ij}^{\lambda,\tau})=\operatorname{sign}\left(\Theta_{ij}^{*}\right), \forall \widetilde{\bTheta}^{\lambda,\tau}_{ij} \in \widetilde{\bTheta}^{\lambda,\tau}\right)   \geq 1  - b_1 \exp(-b_2 n \lambda^2) \to 1
\end{align*}
\end{theorem}
}

Here, $M_{\text{off}}$ refers to all the off-diagonal entries of matrix $M$. Combining Proposition~\ref{proposition_rothman_off_diag} and Assumption~\ref{assumptionA4}, we are able to establish the following graphical model selection consistency result.  We prove Theorem~\ref{theorem1} in Section~\ref{appen_tglasso} of the Supplementary Materials.
\begin{theorem}[restate = theoremone, name = ]
\label{theorem1}
Let assumptions in Proposition~\ref{proposition_rothman_off_diag} hold true. Further, if Assumption~\ref{assumptionA4} holds true with $c_1> 2c_2$, where $c_2$ is defined in Proposition~\ref{proposition_rothman_off_diag}, the thresholded graphical Lasso estimate $\widetilde{\bTheta}^{\lambda,\tau}$ with threshold level $\tau = c_{2}  \sqrt{\frac{s \log p}{n}}$ satisfies:
\begin{align*}
    &\mathbb{P}\left(\operatorname{sign}(\widetilde{\bTheta}_{ij}^{\lambda,\tau})=\operatorname{sign}\left(\Theta_{ij}^{*}\right), \forall \widetilde{\bTheta}^{\lambda,\tau}_{ij} \in \widetilde{\bTheta}^{\lambda,\tau}\right)   \geq 1  - b_1 \exp(-b_2 n \lambda^2) \to 1, \hspace{6mm} \forall i \neq j.
\end{align*}
\end{theorem}


We obtain the same graphical model selection consistency result as shown by \citet{ravikumar2011high}. Yet, more importantly, our result involves less restrictive assumptions on the inverse covariance matrix, i.e., upper and lower bounds on the eigenvalues, whereas the result by \citet{ravikumar2011high} requires strong conditions on the matrix such as irrepresentable or incoherence conditions. In general, such advantage of thresholded estimator applies in many cases, such as in regularized regression \citep{lounici2008sup,meinshausen2009lasso}. Additionally, compared with graphical Lasso which is known to select many false positive edges \citep{lafit2019partial}, thresholded graphical Lasso can zero out such edges. Note there is an extra $\sqrt{s}$ term in the minimum signal strength assumption compared with graphical Lasso. This term cannot be relaxed as we establish Frobenius-norm error bound instead of $\ell_{\infty}$-norm  and $\| \cdot \|_{\infty} \leq\| \cdot \|_{F} \leq \sqrt{s}\| \cdot \|_{\infty}$. Yet, establishing $\ell_{\infty}$-norm error bound  requires irrepresentable condition \citep{ravikumar2011high}. This has also been suggested by \citet{meinshausen2009lasso,lee2015model}.

\subsection{Comparison: Thresholded Graphical Lasso Versus Graphical Lasso with Increasing $\lambda$}\label{glasso_increase_lambda}

One might suggest that the thresholded graphical Lasso is the same as the graphical Lasso with increasing $\lambda$. We find that these two are not equivalent in that the edges estimated by graphical Lasso with increasing $\lambda$ are fundamentally different from thresholded graphical Lasso.  We show empirical evidences in Section~\ref{simulation}.  In terms of theory, as mentioned, graphical Lasso requires irrepresentable condition to ensure graph selection consistency while thresholded graphical Lasso requires that the eigenvalues of the covariance matrices are bounded, a much weaker condition to hold.

The difference between thresholded graphical Lasso and graphical Lasso is reminiscent of the difference between thresholded Lasso and  Lasso:  For Lasso, if there are correlated variables, increasing $\lambda$ does not necessarily zero out noise variables with small coefficients, depending on the correlation structure. To address this, many in the Lasso community proposed thresholded initial Lasso \citep{van2011adaptive,meinshausen2009lasso}.
Recently, \citet{weinstein2020power,wang2020bridge} proved theoretically that thresholded Lasso outperforms Lasso on variable selection and demonstrated rigorous numerical experiments. Particularly, Figure 1 of \citet{weinstein2020power}  suggests that false positives of Lasso cannot be eliminated by increasing $\lambda$, but via thresholding the solution with small $\lambda$. 
We refer readers to Section 3.2 of \citet{su2017false} for further intuition why Lasso does not select some true variables for large $\lambda$.
In brief, Lasso estimates with large $\lambda$ are seriously biased downwards; therefore some null variables may get picked up. 
In this paper,  we show that this also holds for the graphical model case.

\subsection{Extensions: Thresholded Neighborhood Selection and Thresholded CLIME} \label{tnbs}


For CLIME estimator, \citet{cai2011constrained} proposed an additional
thresholding step based on the estimator to yield graphical model selection consistency. 

We investigate the thresholded neighborhood selection and its theoretical properties in Section~\ref{appen_tns} of the Supplementary Materials. In particular, under similar assumptions on the covariance matrix, we show it is graphical model selection consistent. Specifically, we use the result by \citet{lounici2008sup} which requires that the maximum entries of the covariance matrix is upper bounded. Meanwhile, \citet{meinshausen2009lasso}
also proved model selection consistency of thresholded Lasso estimator by establishing $\ell_2$-norm convergence rate which requires incoherent design associated with eigenvalues of the covariance matrix, similar to Assumptions~\ref{assumptionA2} and~\ref{assumptionA3} for the thresholded graphical Lasso case.

For thresholded graphical Lasso, Theorem~\ref{theorem1} requires that the eigenvalues of the inverse covariance matrix has upper and lower bounds. For thresholded neighborhood selection, we have the similar assumption on the covariance matrix. Thresholded CLIME requires the same assumption as CLIME estimator. In terms of sample complexity, graphical Lasso requires sample complexity $n \geq c_2 d^2 \log p$. Thresholded graphical Lasso requires  sample complexity $n \geq c_2 s \log p$. CLIME and thresholded CLIME requires sample complexity $n \geq c_2  \log p$. Neighborhood selection requires $n \geq c_2 d \log p$.


\section{Thresholded Graphical Lasso In the Presence of Latent Variables}
\label{tgraph_lv}

In this section, we study thresholded graphical Lasso in the presence of latent variables and consider conditions when thresholded graphical Lasso can yield a consistent estimate of the sparse concentration matrix $S$ in the latent variable graphical model. In particular, we investigate what properties of the matrix associated with the effects of latent variables, $L$, is required. Also, we investigate the case for thresholded neighborhood selection and thresholded CLIME. Again, all proofs are given in the Supplementary Materials.

\subsection{Thresholded Graphical Lasso In the Presence of Latent Variables}
\label{tglasso_lv}

First, we consider the case for regular graphical Lasso. We discuss the result in detail in Section~\ref{appen_glasso_lv} of the Supplementary Materials. In addition to the assumptions required for graphical Lasso to establish graphical model selection consistency without latent variables, we have the following assumption:
\begin{assumption}[restate = assumptionBfour, name = ]
\label{assumptionB4}
$\|  (S^* - L^*)^{-1} -  (S^*)^{-1}   \|_{\infty}  = \mathcal O\bigg( \sqrt{\frac{\log p}{n}} \bigg)$.
\end{assumption}

Assumption~\ref{assumptionB4} requires that the maximum entries of the quantity $(S^* - L^*)^{-1} -  (S^*)^{-1}$, i.e., the difference between the covariance matrix with and without latent variables, should not exceed the order of $\mathcal O\bigg( \sqrt{\frac{\log p}{n}} \bigg)$. Note this quantity is closely related to the $\| L \|_{\infty}$ by noting that:
\begin{align*}
    (S- L)^{-1} -  S^{-1}  & = (S-L)^{-1} \left[ I - (S-L) S^{-1} \right]  \\
        & =     (S-L)^{-1} \left[ S - (S-L) \right] S^{-1} \\
        & =  (S-L)^{-1} L S^{-1}.
\end{align*}
Since we assume the covariance $(S-L)^{-1}$ satisfies tail bound and $S$ satisfies the minimum signal strength condition, assumption $\| L \|_{\infty} \leq c$ implies that $\| (S- L)^{-1} -  S^{-1} \|_{\infty}  \leq c$. Hence, the quantity $\| (S- L)^{-1} -  S^{-1} \|_{\infty}$ suggests the magnitude of the effect of the latent variables on the covariance matrix.

We show that under irrepresentable condition, minimum signal strength condition and this new assumption,  the graphical Lasso is graphical model selection consistent in the presence of latent variables. We prove Theorem~\ref{theorem2} in Section~\ref{appen_glasso_lv} of the Supplementary Materials.
\hide{
\begin{theorem}
\label{theorem2}
Let irrepresentable condition, minimum signal strength condition (see details in the Supplementary Materials) and Assumption~\ref{assumptionB4} be satisfied. Then, if the sample size $n$ satisfies the bound
$$
n > C_{1} d^{2}\left(1+\frac{8}{\alpha}\right)^{2}(\tau \log p+\log 4)
$$
then with probability greater than $1 - 1/p^{\tau - 2}$, 
the graphical Lasso estimator $\widehat \bTheta^{\lambda}$ is model selection consistent with high probability as $p \to \infty$,
$$
\mathbb{P}\left(\operatorname{sign}(\widehat{\bTheta}_{ij}^{\lambda})=\operatorname{sign}\left(S_{ij}^{*}\right), \forall \widehat{\bTheta}_{ij}^{\lambda} \in \widehat{\bTheta}^{\lambda}\right) \geq 1-1 / p^{\tau-2} \rightarrow 1.
$$
\end{theorem}
}
\begin{theorem}[restate = theoremtwo, name = ]
\label{theorem2}
Let Assumptions~\ref{assumptionB1}-\ref{assumptionB3} and Assumption~\ref{assumptionB4} be satisfied.
Then, if the sample size $n$ satisfies the bound
\begin{align}
n > C_{1} d^{2}\left(1+\frac{12}{\alpha}+C_2\right)^{2}(\tau \log p+\log 4) \label{eq:31},
\end{align}
then with probability greater than $1 - 1/p^{\tau - 2}$, 
the graphical Lasso estimator $\widehat \Theta^{\lambda}$ with regularization parameter $\lambda = (12/\alpha) \bar{\delta}_{f}\left(n, p^{\tau}\right)$ is model selection consistent with high probability as $p \to \infty$,
\begin{align*}
\mathbb{P}\left(\operatorname{sign}(\widehat{\Theta}^{\lambda}_{ij})=\operatorname{sign}\left(S_{ij}^{*}\right), \forall \widehat{\Theta}^{\lambda}_{ij} \in \widehat{\Theta}^{\lambda}\right) \geq 1-1 / p^{\tau-2} \rightarrow 1, 
\end{align*}
\end{theorem}
where $C_1$, $C_2$ and $\bar{\delta}_{f}\left(n, p^{\tau}\right)$ are specified in Section~\ref{appen_glasso_lv} of the Supplementary Materials.

    One might argue that we can directly yield Theorem~\ref{theorem2} with the assumption $\| L^* \|_{\infty} \leq c \sqrt{\frac{\log p}{n}}$ by applying triangular inequality to the $\ell_{\infty}$-norm error bound result of graphical Lasso by \citet{ravikumar2011high}, the proof approach which \citet{ren2012} also used for the CLIME case in the presence of latent variables.  Yet, this thereby requires that $S-L$ satisfies the irrepresentable condition (Assumption 1 by \citet{ravikumar2011high}),  
    which is 
    more restrictive than ours on $S$ since in the original assumption, the irrepresentable condition applies to a sparse matrix.

Similarly, we show that thresholded graphical Lasso can recover the true support of the concentration matrix $S^*$ in the presence of latent variables. First, we establish Frobenius norm convergence  for the  graphical Lasso estimator in the presence of latent variables.

\begin{assumption}[restate = assumptionCone, name = ]
\label{assumptionC1}
$X_{i} \text { be i.i.d. } \mathcal{N}\left(\mathbf{0}, \Sigma^* \right)$ where $\Sigma^* = (S^* - L^*)^{-1}$.
\end{assumption}
\begin{assumption}[restate = assumptionCtwo, name = ]
\label{assumptionC2}
 $\varphi_{\min }\left( {S^*}^{-1} \right) \geq \underline{k}>0,$ or equivalently $\varphi_{\max }\left(S^*\right) \leq 1 / \underline{k}$.
\end{assumption}
\begin{assumption}[restate = assumptionCthree, name = ]
\label{assumptionC3}
$\varphi_{\max }\left({S^*}^{-1}\right) \leq \bar{k}$.
\end{assumption}
\begin{assumption}[restate = assumptionCfour, name = ]
\label{assumptionC4}
Define the minimum signal strength:
\begin{equation*}
\theta_{\min }:=\min _{(i, j) \in E\left(S^{*}\right)}\left|S_{i j}^{*}\right| > c_1 \sqrt{\frac{s \log p}{n}}.
\end{equation*}
\end{assumption}

\begin{lemma}[restate = lemmatwo, name = ]
\label{lemma2}
Let Assumptions~\ref{assumptionB4}-\ref{assumptionC3} be satisfied. If $\lambda \asymp \sqrt{\frac{\log p}{n}}$, there exists a  $c_2$ such that the graphical Lasso estimate $\widehat{\bTheta}^{\lambda}$ satisfies:
$$
\left\| {\widehat{\bTheta}}^{\lambda}-S^{*}\right\|_{F} \leqslant c_{2}  \sqrt{\frac{(p+s) \log p}{n}}   \;\;,
$$
with probability at least $1 - b_1 \exp(-b_2 n \lambda^2)$ where $b_1$ and $b_2$ depend on $\bar{k}$ only.
\end{lemma}

Similarly, as mentioned in Lemma~\ref{lemma1},  the worst part of the rate, $\sqrt{p \log p /n}$, comes from estimating the diagonal. We have:
\begin{proposition}[restate = propositionrothmanoffdiaglv, name = ]
\label{proposition_rothman_off_diag_lv}
Let Assumptions~\ref{assumptionB4}-\ref{assumptionC3} be satisfied. If $\lambda \asymp \sqrt{\frac{\log p}{n}}$, there exists a  $c_2$ such that the graphical Lasso estimate $\widehat{\bTheta}^{\lambda}$ satisfies:
$$
\mathbb{P}\left(  \left\|\widehat{\Theta}_{\text{off}}^{\lambda}-S^{*}_{\text{off}}\right\|_{F} \leq  c_{2}  \sqrt{\frac{s \log p}{n}}  \right)  \geq 1  - b_1 \exp(-b_2 n \lambda^2)  \to 1.
$$
\hide{
$$
\mathbb{P}\left(  \left\|\widehat{\Theta}_{ij}^{\lambda}-S^{*}_{ij}\right\|_{F} \leq  c_{2}  \sqrt{\frac{s \log p}{n}}  \right)  \geq 1  - b_1 \exp(-b_2 n \lambda^2)  \to 1, \hspace{6mm} \forall i \neq j.
$$
}
\end{proposition}

Proof for Proposition~\ref{proposition_rothman_off_diag_lv} suggests that the larger the quantity $\| (S- L)^{-1} -  S^{-1} \|_{\infty}$, the larger the Frobenius-norm error $\left\|\widehat{\Theta}_{\text{off}}^{\lambda}-S^{*}_{\text{off}}\right\|_{F}$. Again, we are able to establish graphical model selection consistency by assuming minimum signal strength condition. 
\hide{
\begin{theorem}
\label{theorem3}
Let Assumption~\ref{assumptionB4}-\ref{assumptionC4}  be satisfied. We assume furthermore that $c_1> 2c_2$, where $c_2$ is defined in Lemma~\ref{lemma2}. Then the thresholded graphical Lasso estimate $\widetilde{\bTheta}^{\lambda,\tau}$ with threshold level $\tau = c_2 \sqrt{\frac{s \log p}{n}}$ satisfies:
\begin{align*}
    &\mathbb{P}\left(\operatorname{sign}(\widetilde{\bTheta}_{ij}^{\lambda,\tau})=\operatorname{sign}\left(S_{ij}^{*}\right), \forall \widetilde{\bTheta}^{\lambda,\tau}_{ij} \in \widetilde{\bTheta}^{\lambda,\tau}\right)  \geq 1  - b_1 \exp(-b_2 n \lambda^2) \to 1.
\end{align*}
\end{theorem}
}

\begin{theorem}[restate = theoremthree, name = ]
\label{theorem3}
Let assumptions in Proposition~\ref{proposition_rothman_off_diag_lv} hold true. Further, if Assumption~\ref{assumptionC4} holds true  with $c_1> 2c_2$, where $c_2$ is defined in 
Proposition~\ref{proposition_rothman_off_diag_lv},
the thresholded graphical Lasso estimate $\widetilde{\bTheta}^{\lambda,\tau}$ with threshold level $\tau = c_2 \sqrt{\frac{s \log p}{n}}$ satisfies:
\begin{align*}
    &\mathbb{P}\left(\operatorname{sign}(\widetilde{\bTheta}_{ij}^{\lambda,\tau})=\operatorname{sign}\left(S_{ij}^{*}\right), \forall \widetilde{\bTheta}^{\lambda,\tau}_{ij} \in \widetilde{\bTheta}^{\lambda,\tau}\right)  \geq 1  - b_1 \exp(-b_2 n \lambda^2) \to 1, \hspace{6mm} \forall i \neq j.
\end{align*}
\end{theorem}

Proofs are given in Section~\ref{appen_tglasso_lv} of the Supplementary Materials. Theorem~\ref{theorem2} and~\ref{theorem3} are novel in comparison to prior analyses as we take the effects of unobserved, latent  variables into account.  In Lemma~\ref{lemma4_rav} and~\ref{lemma6_rav} in the Supplementary Materials, we take into account the quantity $\eta$ associated with latent effects to prove strict dual feasibility and control of deviation. 
We use the proof structure of  \citet{ravikumar2011high}, but well go beyond that to the case of latent variables.  We also include this novel quantity in our proof for Theorem~\ref{theorem3}.

\subsection{Comparison to Latent Variable Graphical Model}


We compare our results as well as assumptions needed with the latent variable graphical model.
First, the latent variable graphical model requires strong irrepresentability or incoherence condition that seems to be difficult to check in practice \citep{ren2012}.  On the other hand, the assumption for our thresholded graphical Lasso estimator involves an upper and lower bound on the eigenvalues of the inverse covariance matrix, which is much less stringent than that of the latent variable graphical model. Moreover, the thresholded graphical Lasso has less sample complexity $n = \mathcal O( s \log p)$ than the latent variable graphical model which requires $n = \mathcal O(p)$.

Also, we compare the assumptions required for the matrix associated with the effects of latent variables, $L^*$, for different methods. Note that although $L$ is termed as the ``low-rank component" by \citet{chandrasekaran2012}, our model does not assume that the matrix $L^*$ has to be low-rank while the latent variable graphical model by \citet{chandrasekaran2012} explicitly assumes that $L^*$ has to be low-rank. Hence our model has a weaker condition on the rank of $L^*$. Further, Assumption~\ref{assumptionB4} requires that the quantity associated with the latent variables, has to vanish, i.e., $\|  (S^* - L^*)^{-1} -  (S^*)^{-1}   \|_{\infty}$ for graphical Lasso and $\| L^* \|_{\infty}$ for CLIME estimator. \citet{chandrasekaran2012} required that the minimum nonzero singular value $\sigma$ of the low-rank matrix $L^*$ has to be greater than $\sqrt{p/n}$. The reason for Assumption~\ref{assumptionB4} is that we want to control the deviation away from the sparse component $S^*$ brought by the latent component,  while \citet{chandrasekaran2012} intended to recover the true rank component by assuming  minimum nonzero singular value. In addition, \citet{chandrasekaran2012} assumed irrepresentability condition associated with the structure between the sparse and low-rank component, an uncheckable and likely stringent condition in practice.

\subsection{Extensions: Thresholded Neighborhood Selection and Thresholded CLIME In the Presence of Latent Variables}

In Section~\ref{appen_tns_lv} of the Supplementary Materials,  we show under similar assumptions on the covariance matrix that the thresholded neighborhood selection is graphical model selection consistent in the presence of latent variables.
Finally, \citet{ren2012} proposed a procedure to obtain an algebraically consistent estimate of the latent variable graphical model based on (thresholded) CLIME estimator. For completeness, we restate the theory for thresholded CLIME in the presence of latent variables in the Supplementary Materials as well. In particular, they required that $\| L^* \|_{\infty} \leq c \sqrt{\frac{\log p}{n}}$, 
a similar assumption we have for thresholded graphical Lasso.

\subsection{Practical Issues}

In this section, we discuss some practical considerations when applying our thresholded graph selection estimators to real data. In particular, we demonstrate approaches to choose the regularization parameter $\lambda$ and the level of threshold $\tau$.

\subsubsection{Choice of $\lambda$}\label{choice_lambda}

Tuning parameter selection for penalized Gaussian graphical models has been well studied in literature. \citet{friedman2008sparse} suggested $K$-fold cross-validation for graphical Lasso. However, cross-validation is known to be liberal for model selection, including many false positives and overfitting the data \citep{wasserman2009high}.  To address this, \citet{foygel2010extended,gao2012tuning} proposed extended Bayesian information criterion (BIC) and demonstrated that the extended BIC yields strong improvement in false discovery rate over the ordinary BIC and more over cross-validation. On the other hand, stability selection for graphical models has also been studied \citep{liu2010stability}. In this paper, we propose using the extended BIC approach since it is computationally faster and works well in practice. We demonstrate the empirical results when the tuning parameter is estimated from the data in Figure~\ref{latent_data_driven} of Section~\ref{simulation}.

\subsubsection{Choice of threshold}\label{choice_thresh}

Similarly, we can adopt the approaches discussed above to choose the optimal combination of regularization $\lambda$ and threshold $\tau$. 

Meanwhile, we find that the edge recovery is robust to the choice of $\lambda$ as long as $\lambda$ is sufficiently small and produces dense solutions. Therefore, we propose to fit regularized graphical model with an initial small $\lambda_0 \propto \sqrt{\frac{\log p}{n}}$ and then i) choose the level of threshold  which gives oracle number of edges or  ii) choose the optimal level of threshold using extended BIC when the oracle number of edges is unknown.  Such approach has also been proposed in the thresholded Lasso literature \citep{zhou2010thresholded,van2011adaptive}. On the other hand, empirical studies show our thresholding graphical model estimator with an initial $\lambda_0$ regularization performs better than thresholding the sample covariance or inverse covariance directly as our approach regularizes the covariance matrix first.

\section{Simulation Studies}
\label{simulation}

In this section, we evaluate the performance of thresholded graphical model estimators by comparing it with the regular graphical model estimators and latent variable graphical model. We consider both the case without latent variables and in the presence of latent variables.

First, we consider the case when there is no latent variable in the model. The conditional graph structure of all the variables is a small world graph with the edge partial correlation coefficients equal to 1. The diagonals of the matrix is the same value $c$ which is chosen so that the covariance matrix is positive definite. We consider both varying $N$ and $p$.

Next we consider the case when there are latent variables in the model. Still, the conditional graphical model structure of all the $p_o$ observed variables are simulated from a small world graph. The $p_h$ hidden variables are fully connected with all $p_o$ observed variables. The entries of the inverse covariance matrix corresponding to the edges between the observed nodes were assigned with value 1, between the observed and the latent variables were assigned with value 0.2, to ensure positive definiteness.  Then, we compute $\widetilde{\bTheta}_{O}^{*}=\left(\Sigma_{O}^{*}\right)^{-1}=\bTheta_{O}^{*}-\bTheta_{O, H}^{*}\left(\bTheta_{H}^{*}\right)^{-1} \bTheta_{H, O}^{*}$ and simulate multivariate normal data from the covariance matrix $\Sigma_{O}^{*}$. Note this data generation process is equivalent to simulating multivariate normal data from $\Sigma^{*} = \bTheta^{-1} = \left(\begin{array}{cc}
        \bTheta_{O} & \bTheta_{O H} \\
        \bTheta_{H O} & \bTheta_{H} \end{array}\right)^{-1}$ and only taking the $p_o$ observed variables, by the construction of latent variable graphical model problem using Schur complement.

Then, we carefully investigate what properties of the graph might affect the quantity $\|  (S^* - L^*)^{-1} -  (S^*)^{-1}   \|_{\infty}$ we have established in Assumption~\ref{assumptionB4} and corresponding Theorem~\ref{theorem2} and \ref{theorem3} (For simplicity, we refer to this quantity as $\eta$.). We  keep sample size fixed $N=150$ and  change one knob of the simulation (properties of the graph) at a time while keeping the rest fixed.  Specifically, we consider the following things to change: number of latent variables, magnitude of connections between the observed and latent variables $\bTheta_{OH}$, magnitude between the latent variables $\bTheta_H$, sparsity level of the connections between the observed and latent variables $\bTheta_{OH}$ and sparsity level between latent variables $\bTheta_H$. To change the magnitude, we multiply the entries of connections in the base simulation by a constant $c$. To change the sparsity, we randomly impose some entries of connections to be zero. Hence, level of sparsity refers to the percentage of zero entries; a greater level of sparsity implies more zero entries in the matrix.

\begin{figure}[h]
\vskip 0.2in
\begin{center}
\centerline{\includegraphics[scale = 0.6]{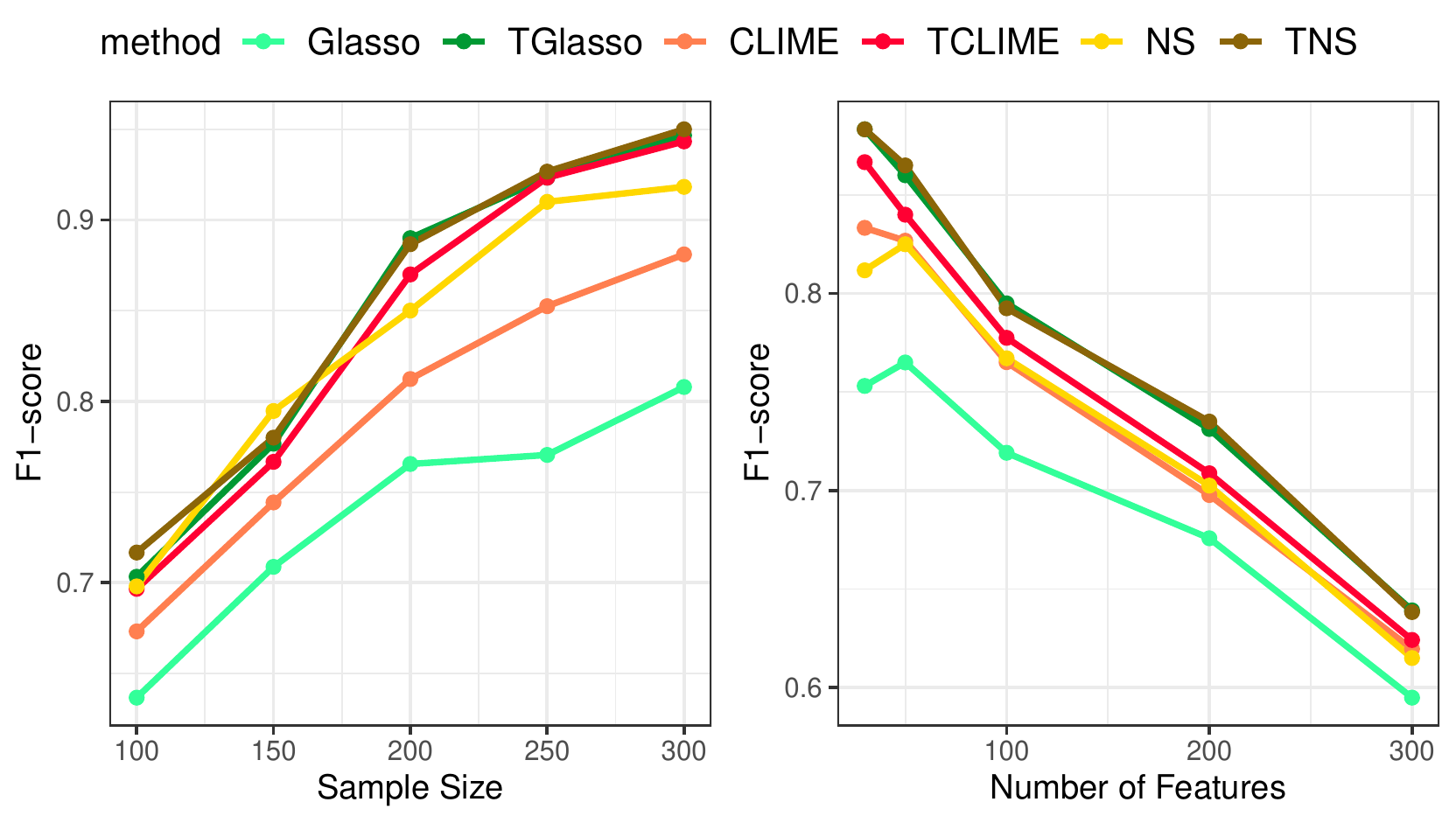}}
\caption{Simulation results for the Gaussian graphical model without latent variables. We consider a small world graph with varying number of samples while $p_o = 30$ (left) and varying features while $N = 150$ (right). We plot the F1-score of edge recovery.}
\label{no_latent}
\end{center}
\vskip -0.2in
\end{figure}

\begin{figure}[h!]
\vskip 0.2in
\begin{center}
\centerline{\includegraphics[scale = 0.8]{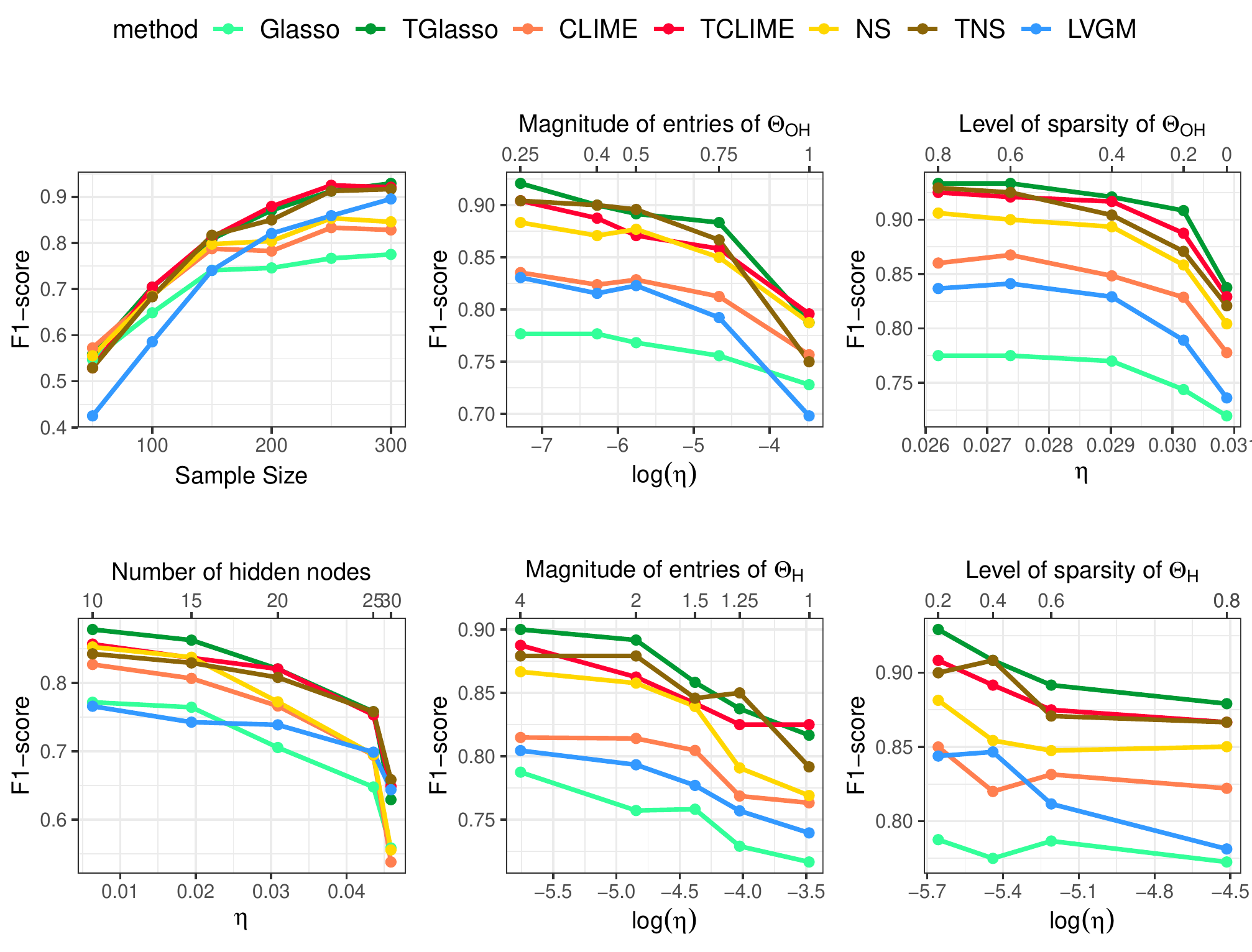}}
\caption{Simulation results for the Gaussian graphical model in the presence of latent variables. We consider a small world graph. For the base simulation (top left plot), the number of observed nodes are set to be 30 and hidden nodes are 20. The edge weights between observed nodes are set to be 1. Hidden nodes are densely connected with the observed nodes with $\bTheta_{OH} = 0.2$. We vary sample sizes in the base simulation. For the rest of the experiments, we keep sample size $N=150$ and change one knob of the simulation setup at a time. We plot F1-score of edge recovery over sample size (top left plot) or plot F1-score over the knobs we change along with the corresponding $\eta$ (rest of the plot). The term $\eta$ refers to $\|  (S^* - L^*)^{-1} -  (S^*)^{-1}   \|_{\infty}$ in Assumption~\ref{assumptionB4}.  Results reveal that thresholded graphical model estimators outperform the latent variable graphical model estimator under nearly all experimental settings.
}
\label{latent_base_sim}
\end{center}
\vskip -0.2in
\end{figure}

\begin{figure}[h!]
\vskip 0.2in
\begin{center}
\centerline{\includegraphics[scale = 0.6]{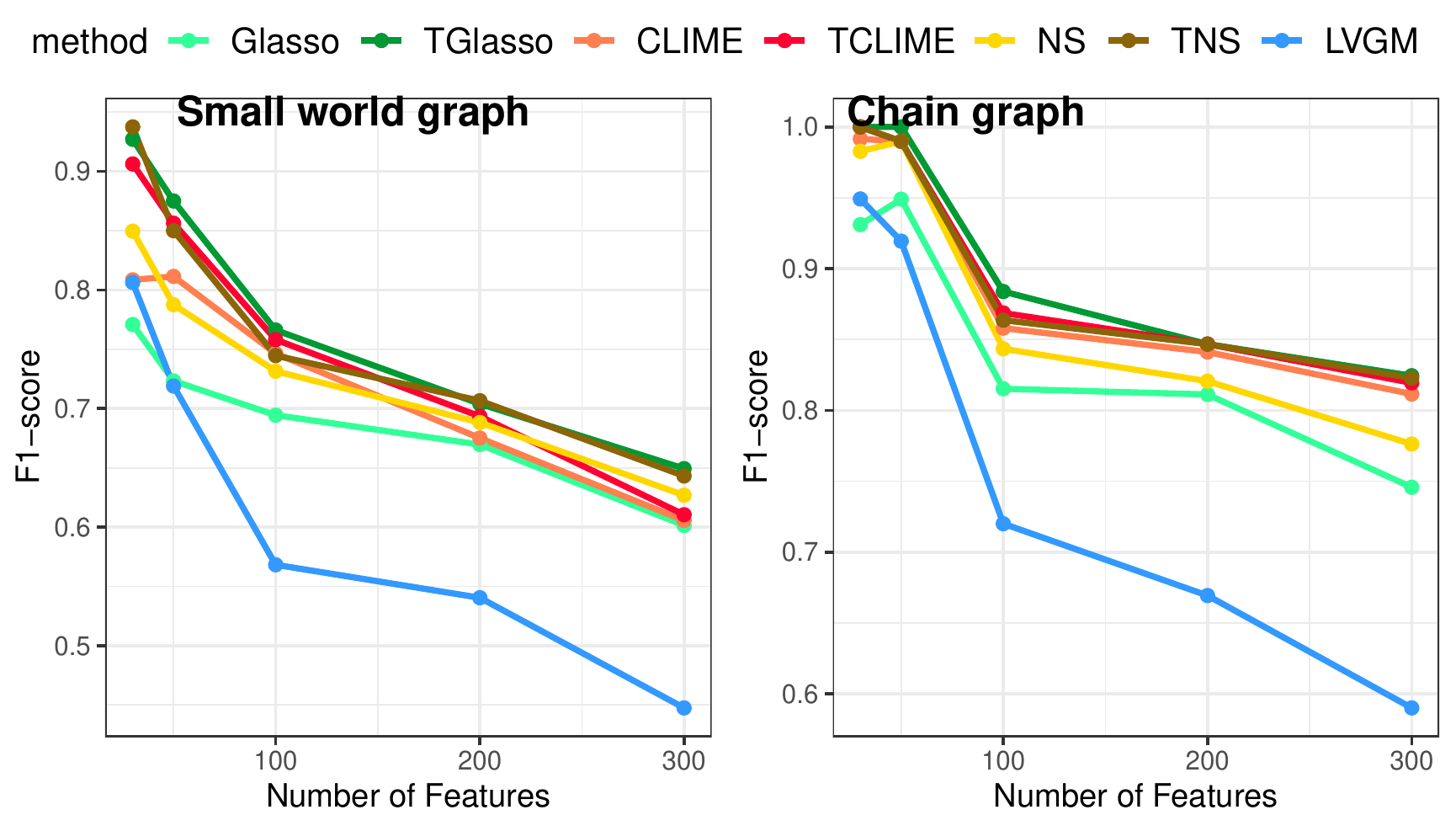}}
\caption{Simulation results for the Gaussian graphical model in high dimensions and in the presence of latent variables. We keep the sample size  at 150 and vary the number of features. We consider different graph structures: a small world graph and a chain graph. The rest of the setup is the same as the base simulation.}
\label{latent_hd}
\end{center}
\vskip -0.2in
\end{figure}

\begin{figure}[ht!]
\vskip 0.2in
\begin{center}
\centerline{\includegraphics[scale = 0.7]{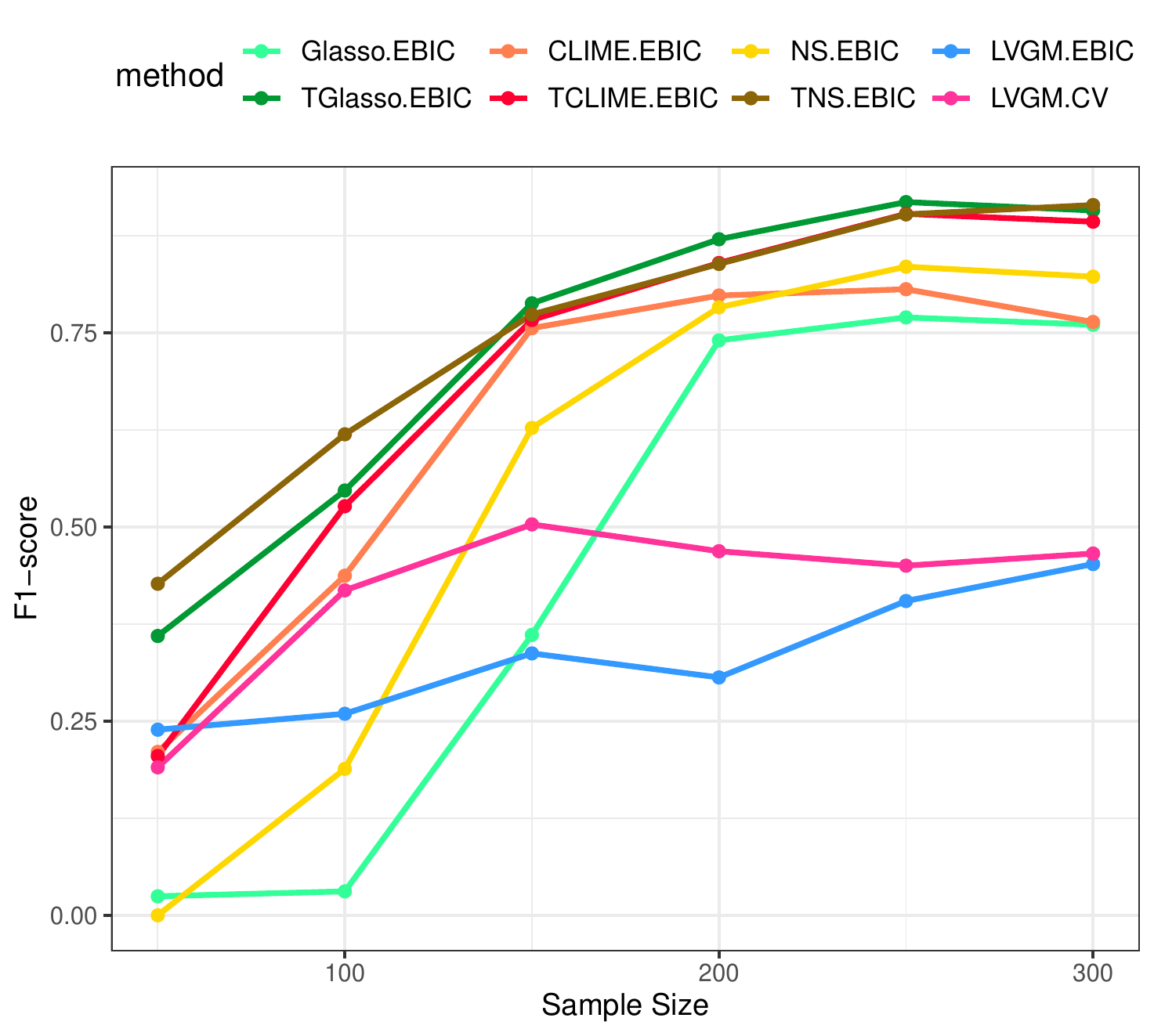}}
\caption{Simulation results for the Gaussian graphical model in the presence of latent variables where estimates used data-driven tuning parameters selection. We consider the same setup as the base simulation (the top left plot in Figure~\ref{latent_base_sim}).}
\label{latent_data_driven}
\end{center}
\vskip -0.2in
\end{figure}

For our simulation results in Figure~\ref{no_latent}, \ref{latent_base_sim} and~\ref{latent_hd}, we use oracle sparsity for all methods (i.e., every method uses tuning parameters that yield the true number of edges), for fair comparisons. 
This means, for graphical Lasso, we choose a (large) $\lambda$ which gives oracle true number of edges. For thresholded graphical Lasso, we also choose a pair of proper regularization $\lambda$ and level of threshold $\tau$ that gives the oracle true number of edges. In general, we find that edge recovery is robust to the choice of $\lambda$ when the level of threshold is then chosen to give oracle number of edges, as long as $\lambda$ is sufficiently small and produces dense solutions, as shown in Figure~\ref{heatmap} in the Supplementary Materials. 
Hence, following \citet{zhou2010thresholded,van2011adaptive}'s approach, we propose to first fit regularized graphical model with an initial (small) $\lambda_0 \propto \sqrt{\frac{\log p}{n}}$, and then choose the threshold level which gives oracle true number of edges. In this way, all methods yield the same number of edges.  From the simulation results, we see that even with optimally tuned $\lambda$ which selects the true number of edges, the graphical Lasso still does not perform well in terms of edge recovery whereas thresholded graphical Lasso does. This suggests that the edges estimated by graphical Lasso with increasing $\lambda$ are fundamentally different from thresholded graphical Lasso, as mentioned in Section~\ref{glasso_increase_lambda}. 
Typically, thresholding a dense solution is better than just using a sparse solution, as also shown in Figure~\ref{heatmap} in the Supplementary Materials. Note we include the results when the number of edges is estimated from the data in Figure~\ref{latent_data_driven}.

For all the results, we run the experiment with 5 replicates, except for the high-dimensional case where we run 2 replicates. To evaluate the edge recovery accuracy of different methods, we use F1-score as the metric. F1-score, a measure of a model's accuracy, is defined as the  harmonic mean of precision and recall, or $F_1 =  \frac{2}{\text { recall }^{-1}+\text {precision }^{-1}} =\frac{\text { TP }}{ \text{TP} +\frac{1}{2}( \text{FP} +  \text{FN})}$. We compute the F1-score by comparing the set of selected edges and set of true edges; hence it measures how closely we capture the true edges of the graph. The F1-scores are averaged over replicates.

Both Figure~\ref{no_latent} and~\ref{latent_base_sim} suggest that thresholding basically improves the estimates. Further, Figure~\ref{latent_base_sim} shows that the term in Assumption~\ref{assumptionB4} and the theorem, $\eta = \|  (S^* - L^*)^{-1} -  (S^*)^{-1}   \|_{\infty}$, affects edge recovery accuracy.  Although Assumption~\ref{assumptionB4} cannot be checked in practice (just like irrepresentable condition), it is more interpretable than the assumptions of \citet{chandrasekaran2012}. In Figure~\ref{latent_base_sim},  we show how the quantity $\eta$ in Assumption~\ref{assumptionB4} changes with respect to different properties of the graph which include the number of hidden nodes, magnitude of connections, and etc; all of these quantities are interpretable. For example, we find that increasing the number of latent variables, the magnitude of the entries of $\bTheta_{OH}$, and the proportion of non-zero entries of $\bTheta_{OH}$ leads to an increase in the quantity $\eta$, and hence worse edge recovery accuracy. This makes sense as $\eta$ is closely related to $\| L \|_{\infty}$ and $L = \bTheta_{O, H}^{*}\left(\bTheta_{H}^{*}\right)^{-1} \bTheta_{H, O}^{*}$.  We have similar results for $\bTheta_H$, but in the opposite manner due to the inverse. Our findings align with Theorem~\ref{theorem2} and \ref{theorem3} which suggest that larger $\eta$, i.e., stronger effects associated with latent variables, leads to worse edge recovery. Also, Figure~\ref{latent_base_sim} suggests that our method works well when $L$ is not low-rank.

We also consider the high-dimensional setting in the presence of latent variables with different graph structures. 
Figure~\ref{latent_hd} suggests that our thresholded estimators outperform the latent variable graphical model when $p>n$.

Finally, we estimate the tuning parameters $\lambda$ and threshold level based on the data using the approaches proposed in Section~\ref{choice_thresh} and show the results in Figure~\ref{latent_data_driven}. As discussed in Section~\ref{choice_lambda}, we find that extended BIC outperforms cross-validation approach. We also extend the extended BIC for latent variable graphical model. For fair comparisons, we also show the results using cross-validation for latent variable graphical model. We perform 5-fold cross validation. Figure~\ref{latent_data_driven} suggests that our proposed data-driven tuning parameter selection approach using EBIC works well in practice and thresholded estimators still outperform the original estimator and the latent variable graphical model approach.

\section{Case Study: Functional Neural Connectivity}
\label{realdata}

We evaluate the performance of our method on a publicly available calcium imaging data from the Allen Brain Atlas \citep{lein2007genome}. The data set contains fluorescence traces of simultaneously recorded neurons in the mouse visual cortex during spontaneous neural activity under various visual stimuli. We analyze neural responses during one type of visual stimuli: drifting angular gratings.  The data during this stimulus consists of 115,735 time points for 227 neurons.

\begin{figure*}[ht]
\vskip 0.2in
\begin{center}
\centerline{\includegraphics[scale = 0.35]{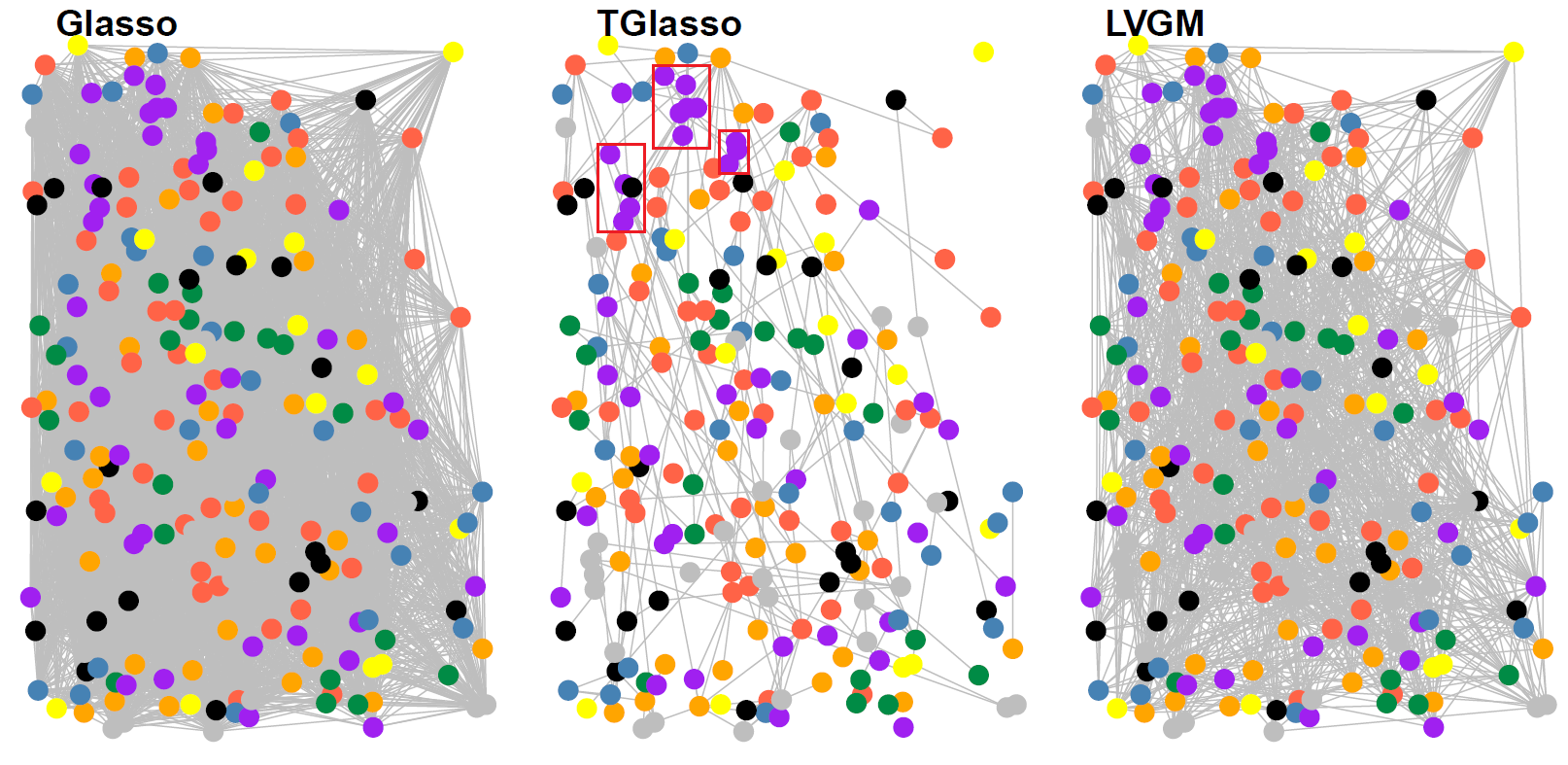}}
\caption{Graphical model estimates of functional neural connectivity from calcium imaging data in the visual cortex. The placement of nodes in the graphs reflects neuron locations and the nodes are colored according to their neural tuning. Our thresholded graphical Lasso (middle) identifies key connections between similarly tuned neurons.}
\label{real_data}
\end{center}
\vskip -0.2in
\end{figure*}

We fit the graphical Lasso, our thresholded graphical Lasso and the latent variable graphical model to this data with results shown in  Figure~\ref{real_data}. We see that the graphical Lasso produces a very dense graph as expected since there are large numbers of latent unobserved neurons.  The latent variable graphical model estimate is less dense, but still denser than expected for functional neural connections which typically follow a small-world structure.  On the other hand, our thresholded graphical Lasso approach identifies a reasonably sparse graph by eliminating small, and likely false positive edges, through thresholding.

While there is no ground truth by which we can evaluate our graphical model estimates, we examine how the graph estimates relate to neural tuning.  Neural tuning refers to neurons in the visual cortex that selectively activate in response to specific visual angles.  Neuroscientists have suggested that neurons that are tuned similarly are more likely to be functionally connected \citep{yatsenko2015improved}.  In this experiment, drifting gratings stimuli was presented at various angular frequencies and for eight different angular modalities at multiples of 45 degrees.  The neural tuning was measured using the global orientation selectivity index.  In Figure~\ref{real_data}, we color the neurons according to their neural tuning to the eight different angles. Interestingly, we find that our thresholded graphical Lasso is able to identify connections between hubs that have neurons with the same neuron tuning information (highlighted in red boxes).  For a quantitative comparison, we show the proportion of edges that share the same neural tuning in Table~\ref{real-data-table}. The thresholded graphical lasso seems to more often connect neurons that have the same neural tuning, thus validating our approach.

\begin{table}[ht!]
\caption{Proportion of edges between two neurons that share the same neural tuning}
\label{real-data-table}
\vskip 0.15in
\begin{center}
\begin{small}
\begin{sc}
\begin{tabular}{lcccr}
\toprule
Method & Glasso & TGlasso & LVGM \\
\midrule
Proportion    & 18\%  & 32\% &  29\% \\
\bottomrule
\end{tabular}
\end{sc}
\end{small}
\end{center}
\vskip -0.1in
\end{table}

\section{Conclusion}
\label{conclusion}
In this paper, we propose a simple solution to solve the graph selection in the presence of latent variables: apply a hard thresholding operator to existing graph selection methods. We show that this thresholding approach is not only computationally faster than the convex program of the latent variable graphical model, but it also enjoys better theoretical properties and has superior performance in practice.  In particular, our theoretical analysis reveals that thresholded graphical model estimates are graph selection consistent under much weaker assumptions and lower sample complexity than that of the latent variable graphical model.  Additionally, our method shows promise as a tool to estimate functional neural connectivity in the presence of large numbers of unobserved or latent neurons.
We propose a simple, general thresholding framework that can be easily extended to other graph estimators such as Ising models and other exponential family graphical models \citep{yang2015graphical}.  On the other hand, we study undirected Gaussian graphical models.  It will be interesting to investigate if thresholding can yield the same graph selection consistency  for the directed acyclic graphs (DAG) case in the presence of latent variables.
Overall, our work suggests that simple thresholding solutions may be superior both statistically and computationally for the task of graph selection in the presence of latent variables.


\section*{Acknowledgements}

The authors acknowledge support from NSF DMS-1554821, NSF NeuroNex-1707400, and NIH 1R01GM140468.
The  authors  thank  Andersen Chang and Dr. Kre\v{s}imir Josi\'c 
for helpful discussions on the calcium imaging data.

\newpage
\begin{appendix}

\begin{center}
{\bf \LARGE Thresholded Graphical Lasso Adjusts for Latent Variables: Application to Functional Neural Connectivity: Supplementary Materials}
\bigskip

{\large Minjie Wang and Genevera I. Allen}
\end{center}


\section{Thresholded Graphical Lasso}\label{appen_tglasso}
 
In this section, we study the theoretical properties of thresholded graphical Lasso in terms of graphical model selection consistency for the Gaussian graphical model.  We show graph selection consistency based on the results of \citet{rothman2008sparse}, who established convergence in Frobenius norm for graphical Lasso estimator. We make the following assumptions on our model. Denote $\varphi_{\max}(\cdot)$ and $\varphi_{\min}(\cdot)$ as the largest and smallest eigenvalues of a matrix. Denote $s$ as the total number of non-zero edges, i.e., $E\left(\Theta^{*}\right):=\left\{ (i, j) \in V \times V \mid i \neq j, \Theta_{i j}^{*} \neq 0\right\}$ and $s = |E \left(\Theta^{*}\right)|$. 

\hide{
\begin{assumption}
\label{assumptionA1}
$X_{i} \text { be i.i.d. } \mathcal{N}\left(\mathbf{0}, \Sigma^* \right)$.
\end{assumption}
\begin{assumption}
\label{assumptionA2}
$\varphi_{\min }\left(\Sigma^*\right) \geq \underline{k}>0,$ or equivalently $\varphi_{\max }\left(\Theta^*\right) \leq 1 / \underline{k}$.
\end{assumption}
\begin{assumption}
\label{assumptionA3}
 $\varphi_{\max }\left(\Sigma^*\right) \leq \bar{k}$.
\end{assumption}
}

\assumptionAone*
\assumptionAtwo*
\assumptionAthree*

\hide{
\begin{lemma}[\citealt{rothman2008sparse}]
\label{lemma1}
Let Assumptions~\ref{assumptionA1}-\ref{assumptionA3} be satisfied. If $\lambda \asymp \sqrt{\frac{\log p}{n}}$, there exists a  $\hat c_2$ such that the graphical Lasso estimate $\widehat{\Theta}^{\lambda}$ satisfies:
$$
\mathbb{P}\left(  \left\|\widehat{\Theta}^{\lambda}-\Theta^{*}\right\|_{F} \leq \hat c_{2}  \sqrt{\frac{(p+s) \log p}{n}}  \right) \to 1.
$$
\end{lemma}
}
\lemmaone*

Note, as discussed by \citet{rothman2008sparse}, the worst part of the rate, $\sqrt{p \log p /n}$, comes from estimating the diagonal. Since we are interested in edge recovery, it can be shown that we can get the rate of $\sqrt{s \log p /n}$ for off-diagonal parts by assuming the set $
\Theta_{n}(M)=\left\{\Delta: \Delta=\Delta^{T},\|\Delta^-\|_{F}=M \sqrt{\frac{s \log p}{n}}, \|\Delta^+\|_{F}=M \sqrt{\frac{(p+s) \log p}{n}}  \right\}
$ in the original proof.
Or we could use the correlation matrix rather than the covariance matrix. Hence, we have:
\hide{
$$
\mathbb{P}\left(  \left\|\hat{\Theta}_{\text{off}}^{\lambda}-\Theta^{*}_{\text{off}}\right\|_{F} \leq  c_{2}  \sqrt{\frac{s \log p}{n}}  \right) \to 1.
$$
}

\propositionrothmanoffdiag*

\noindent \underline{\textbf{Proof of Proposition~\ref{proposition_rothman_off_diag}:}} 
We follow closely the proof approach used by \citet{rothman2008sparse}; we consider the deviation of the diagonal and off-diagonal parts separately.
Denote $M^+ = \text{diag}(M)$ for a diagonal matrix with the same diagonal as $M$ and $M^- = M - M^+$. Let
\begin{align}
Q(\bTheta)=& \operatorname{tr}(\bTheta \hat{\Sigma})-\log |\bTheta|+\lambda\left|\bTheta^{-}\right|_{1}-\operatorname{tr}\left(\bTheta^* \hat{\Sigma}\right)+\log \left|\bTheta^*\right|-\lambda\left| {\bTheta^*}^{-}\right|_{1} \nonumber \\
=& \operatorname{tr}\left[\left(\bTheta-\bTheta^*\right)\left(\hat{\Sigma}-\Sigma^{*}\right)\right]-\left(\log |\bTheta|-\log \left|\bTheta^*\right|\right) \nonumber \\
&+\operatorname{tr}\left[\left(\bTheta-\bTheta^*\right) \Sigma^{*}\right]+\lambda\left(\left|\bTheta^{-}\right|_{1}-\left|{\bTheta^*}^{-}\right|_{1}\right) \label{eq:off-diag-7}.
\end{align}

Our estimate $\widehat\bTheta$ minimizes $Q(\bTheta)$, or equivalently $\widehat{\Delta}=\widehat{\bTheta}-\bTheta^*$ minimizes $G(\Delta) \equiv$ $Q\left(\bTheta^*+\Delta\right) .$ 
Consider the set
\hide{
$$
\Theta_{n}(M)=\left\{\Delta: \Delta=\Delta^{T},\|\Delta^+\|_{F}=M^+ \sqrt{\frac{(p+s) \log p}{n}},  \|\Delta^-\|_{F}=M^- \sqrt{\frac{s \log p}{n}}               \right\},
$$
}
$$
\Theta_{n}(M)=\left\{\Delta: \Delta=\Delta^{T},\|\Delta^+\|_{F}=M^+  r_n,  \|\Delta^-\|_{F}=M^-  r'_n           \right\},
$$
where
$$
r_{n}=\sqrt{\frac{(p+s) \log p}{n}} \rightarrow 0, \text{ and } r'_{n}=\sqrt{\frac{s \log p}{n}} \rightarrow 0.
$$
Note that $G(\Delta)=Q\left(\bTheta^*+\Delta\right)$ is a convex function, and
$$
G(\widehat{\Delta}) \leq G(0)=0.
$$
Then, if we can show that
$$
\inf \left\{G(\Delta): \Delta \in \Theta_{n}(M)\right\}>0,
$$
the minimizer $\widehat{\Delta}$ must be inside the sphere defined by $\Theta_{n}(M),$ and hence
\hide{
$$
\|\hat{\Delta}^+ \|_{F} \leq M^+ \sqrt{\frac{(p+s) \log p}{n}},  \text{ and }
\|\hat{\Delta}^- \|_{F} \leq M^- \sqrt{\frac{s \log p}{n}}. $$
}
$$
\|\widehat{\Delta}^+ \|_{F} \leq M^+ r_n,  \text{ and }
\|\widehat{\Delta}^- \|_{F} \leq M^- r_n'. $$

For the logarithm term in \eqref{eq:off-diag-7}, we use Taylor expansion of $f(t)=\log |\bTheta+t \Delta|$ and integral form of the remainder:
$$
\log \left|\bTheta^* +\Delta\right|-\log \left|\bTheta^*\right|=\operatorname{tr}\left(\Sigma^* \Delta\right)-\tilde{\Delta}^{T}\left[\int_{0}^{1}(1-v)\left(\bTheta^*+v \Delta\right)^{-1} \otimes\left(\bTheta^*+v \Delta\right)^{-1} d v\right] \tilde{\Delta},
$$
where $\otimes$ is the Kronecker product 
and $\tilde{\Delta}$ is $\Delta$ vectorized to match the dimensions of the Kronecker product.

Therefore, we may write \eqref{eq:off-diag-7} as,
\begin{align}
G(\Delta)=& \operatorname{tr}\left(\Delta\left(\hat{\Sigma}-\Sigma^* \right)\right)+\tilde{\Delta}^{T}\left[\int_{0}^{1}(1-v)\left(\bTheta^*+v \Delta\right)^{-1} \otimes\left(\bTheta^*+v \Delta\right)^{-1} d v\right] \tilde{\Delta} \nonumber \\
&+\lambda\left(\left| {\bTheta^*}^{-}+\Delta^{-}\right|_{1}-\left|{\bTheta^*}^{-}\right|_{1}\right) \label{eq:off-diag-10}.
\end{align}

For an index set $A$ and a matrix $M=\left[m_{i j}\right]$, we denote $M_{A} = \left[m_{i j} I((i, j) \in A)\right]$, where $I(\cdot)$ is an indicator function. Recall $S=\left\{(i, j): \bTheta^*_{i j} \neq 0, i \neq j\right\}$ and denote $\overline{S}$ as its complement. Note that $\left| {\bTheta^*}^{-}+\Delta^{-}\right|_{1}=\left| {\bTheta^*_{S}}^{-}+\Delta_{S}^{-}\right|_{1}+\left|\Delta_{\overline{S}}^{-}\right|_{1},$ and
$\left| {\bTheta^*}^{-}\right|_{1}=\left| {\bTheta^*_{S}}^{-}\right|_{1}$. Then the triangular inequality implies
$$
\lambda\left(\left| {\bTheta^*}^{-}+\Delta^{-}\right|_{1}-\left| {\bTheta^*}^{-}\right|_{1}\right) \geq \lambda\left(\left|\Delta_{\overline{S}}\right|_{1}-\left|\Delta_{S}^{-}\right|_{1}\right).
$$

Again, by triangular inequality, we have
$$
\left|\operatorname{tr}\left(\Delta\left(\hat{\Sigma}-\Sigma^* \right)\right)\right| \leq\left|\sum_{i \neq j}\left(\hat{\sigma}_{i j}-\sigma_{0 i j}\right) \Delta_{i j}\right|+\left|\sum_{i}\left(\hat{\sigma}_{i i}-\sigma_{0 i i}\right) \Delta_{i i}\right|=\mathrm{I}+\mathrm{II}.
$$

To bound term I, note that the union sum inequality and  Lemma 1 of \citet{rothman2008sparse} (also Lemma 3 of \citet{bickel2008regularized}) imply that, with probability tending to 1,
$$
\max _{i \neq j}\left|\hat{\sigma}_{i j}-\sigma_{0 i j}\right| \leq C_{1} \sqrt{\frac{\log p}{n}},
$$
and hence term I is bounded by
$$
\mathrm{I} \leq C_{1} \sqrt{\frac{\log p}{n}}\left|\Delta^{-}\right|_{1}.
$$

The second bound comes from the Cauchy-Schwartz inequality and Lemma 1 of \citet{rothman2008sparse}:
\begin{align*}
\mathrm{II} & \leq\left[\sum_{i=1}^{p}\left(\hat{\sigma}_{i i}-\sigma_{i i}\right)^{2}\right]^{1 / 2}\left\|\Delta^{+}\right\|_{F} \leq \sqrt{p} \max _{1 \leq i \leq p}\left|\hat{\sigma}_{i i}-\sigma_{0 i i}\right|\left\|\Delta^{+}\right\|_{F} \\
    & \leq C_2 \sqrt{\frac{p \log p}{n}}\left\|\Delta^{+}\right\|_{F} \leq C_2 \sqrt{\frac{(p+s) \log p}{n}}\left\|\Delta^{+}\right\|_{F} ,
\end{align*}
also with probability tending to 1.

Now, take
$$
\lambda=\frac{C_{1}}{\varepsilon} \sqrt{\frac{\log p}{n}}.
$$
By~\eqref{eq:off-diag-10},
$$
\begin{aligned}
G(\Delta) \geq & \frac{1}{4} \underline{k}^{2}\|\Delta\|_{F}^{2}-C_{1} \sqrt{\frac{\log p}{n}}\left|\Delta^{-}\right|_{1}-C_{2} \sqrt{\frac{(p+s) \log p}{n}}\left\|\Delta^{+}\right\|_{F} \\
&+\lambda\left(\left|\Delta_{\overline{S}}^{-} \right|_{1}-\left|\Delta_{S}^{-}\right|_{1}\right) \\
=& \frac{1}{4} \underline{k}^{2}\|\Delta\|_{F}^{2}-C_{1} \sqrt{\frac{\log p}{n}}\left(1-\frac{1}{\varepsilon}\right)\left|\Delta_{\overline{S}}^{-} \right|_{1}-C_{1} \sqrt{\frac{\log p}{n}}\left(1+\frac{1}{\varepsilon}\right)\left|\Delta_{S}^{-}\right|_{1} \\
&-C_{2} \sqrt{\frac{(p+s) \log p}{n}}\left\|\Delta^{+}\right\|_{F}.
\end{aligned}
$$
The first term comes from a bound on the integral which we will show below. The second term is always positive given small $\epsilon$, and hence we may omit it for the lower bound. Now, note that
$$
\left|\Delta_{S}^{-}\right|_{1} \leq \sqrt{s}\left\|\Delta_{S}^{-}\right\|_{F} \leq \sqrt{s}\left\|\Delta^{-}\right\|_{F} \leq \sqrt{p+s}\left\|\Delta^{-}\right\|_{F}.
$$

Thus we have
$$
\begin{aligned}
G(\Delta) & \geq\left\|\Delta^{-}\right\|_{F}^{2}\left[\frac{1}{4} \underline{k}^{2}-C_{1} \sqrt{\frac{s \log p}{n}}\left(1+\frac{1}{\varepsilon}\right)\left\|\Delta^{-}\right\|_{F}^{-1}\right] \\
&+\left\|\Delta^{+}\right\|_{F}^{2}\left[\frac{1}{4} \underline{k}^{2}-C_{2} \sqrt{\frac{(p+s) \log p}{n}}\left\|\Delta^{+}\right\|_{F}^{-1}\right] \\
&=\left\|\Delta^{-}\right\|_{F}^{2}\left[\frac{1}{4} \underline{k}^{2}-\frac{C_{1}(1+\varepsilon)}{\varepsilon M^-}\right]+\left\|\Delta^{+}\right\|_{F}^{2}\left[\frac{1}{4} \underline{k}^{2}-\frac{C_{2}}{M^+}\right]>0,
\end{aligned}
$$
for $M^+$ and $M^-$ sufficiently large. 

It only remains to check the bound on the integral term in~\eqref{eq:off-diag-10}. Recall that $\varphi_{\min }(W)=\min _{\|x\|=1} x^{T} W x .$ After factoring out the norm of $\tilde{\Delta},$ we have, for $\Delta \in \Theta_{n}(M)$,
$$
\begin{aligned}
\varphi_{\min } &\left(\int_{0}^{1}(1-v)\left( \bTheta^* +v \Delta\right)^{-1} \otimes\left( \bTheta^*+v \Delta\right)^{-1} d v\right) \\
& \geq \int_{0}^{1}(1-v) \varphi_{\min }^{2}\left( \bTheta^* +v \Delta\right)^{-1} d v \geq \frac{1}{2} \min _{0 \leq v \leq 1} \varphi_{\min }^{2}\left( \bTheta^* +v \Delta\right)^{-1} \\
& \geq \frac{1}{2} \min \left\{\varphi_{\min }^{2}\left( \bTheta^* +\Delta\right)^{-1}: \|\Delta^+\|_{F} \leq M^+ \sqrt{\frac{(p+s) \log p}{n}},  \|\Delta^-\|_{F} \leq M^- \sqrt{\frac{s \log p}{n}}  \right\}.
\end{aligned}
$$
The first inequality holds true since the eigenvalues of the Kronecker products of symmetric matrices are equal to the products of the eigenvalues of their factors. Now
$$
\varphi_{\min }^{2}\left( \bTheta^* +\Delta\right)^{-1}=\varphi_{\max }^{-2}\left( \bTheta^*+\Delta\right) \geq\left(\left\|\bTheta^* \right\|+\|\Delta\|\right)^{-2} \geq \frac{1}{2} \ubar{k^{2}},
$$
with probability tending to $1,$ since $\|\Delta\| \leq\|\Delta\|_{F}=o(1) .$ This establishes the theorem.
$\qed$

To establish graphical model selection consistency, we further assume the minimum signal strength condition Assumption~\ref{assumptionA4}, an assumption usually stated for model selection consistency.

\hide{
\begin{assumption}
\label{assumptionA4}
Define the minimum signal strength:
\begin{equation*}
 \theta_{\min }:=\min _{(i, j) \in E\left(\Theta^{*}\right)}\left|\Theta_{i j}^{*}\right| > c_1 \sqrt{\frac{s \log p}{n}}.   
\end{equation*}
\end{assumption}
}
\assumptionAfour*

\theoremone*

\noindent \underline{\textbf{Proof of Theorem~\ref{theorem1}:}} 

Denote $E(\bTheta^*)$ as the true edge set, i.e., $E(\bTheta^*) = \left\{ (i, j) \in V \times V   \mid   \Theta_{i j}^{*} \neq 0, i \neq j \right\}$. 

For $(i,j) \in E^C$, we have $\bTheta_{ij}^* = 0$ and from 
Proposition~\ref{proposition_rothman_off_diag}, $|\widehat \bTheta^{\lambda}_{ij}| < c_2 \sqrt{\frac{s \log p}{n}}$. Then by the definition of thresholded graphical Lasso estimator, $\widetilde \bTheta^{\lambda,\tau}_{ij} = 0$.

For $(i,j) \in E$, by minimum signal strength condition, we have $|\bTheta^*_{ij}| \geq c_1 \sqrt{\frac{s \log p}{n}}$ and from 
Proposition~\ref{proposition_rothman_off_diag}, $\left| \widehat \bTheta^{\lambda}_{ij} - \bTheta_{ij}^*\right| < c_2 \sqrt{\frac{s \log p}{n}}$. Note we assume $c_1 > 2 c_2$. Therefore, $|\widehat \bTheta^{\lambda}_{ij}| > c_2 \sqrt{\frac{s \log p}{n}}$. Then by the definition of thresholded graphical Lasso estimator, $\widetilde \bTheta^{\lambda,\tau}_{ij} > 0$.  $\qed$

\section{Thresholded Neighborhood Selection}\label{appen_tns}
To study the theoretical properties of thresholded neighborhood selection, we use the results of \citet{lounici2008sup} as building block. First, we have the following assumptions.
\begin{assumption}
\label{assumptionNB1}
The random variables $\epsilon_{1},\cdots,\epsilon_{n}$ are i.i.d. $\mathcal{N}\left(0, \sigma^{2}\right)$.
We also need assumptions on the Gram matrix:
$$
\Psi \triangleq \frac{1}{n} X^{T} X.
$$
\end{assumption}
\begin{assumption}
\label{assumptionNB2}
The elements $\Psi_{i, j}$ of the Gram matrix $\Psi$ satisfy
$$
\Psi_{j, j}=1, \quad \forall 1 \leq j \leq p
$$
and
$$
\max_{ 1\leq k \leq p}\max _{i,j \in \{1,\cdots,p\} \backslash k,  i \neq j}\left|\Psi_{i, j}\right| \leq \frac{1}{\alpha\left(1+2 c_{0}\right) s}  \;,
$$
where $c_0 = 3$ as we consider the Lasso estimator.
\end{assumption}

Following the notation in Chapter 12 of \citet{maathuis2018handbook}, we define the $j^{\text{th}}$ nodewise regression coefficient $\theta_j$ as the solution to the Lasso program:
\[
\widehat{\theta}_{j}=\argmin_{\theta}\left\{\left\|X^{j}-X^{	\setminus j} \theta\right\|_{2}^{2}+\lambda_{j}\|\theta\|_{1}\right\},
\]
where $X^{j}$ denotes the $j^{\text {th }}$ column vector of the $n \times p$ data matrix and $X^{\setminus j}$ denotes the $n \times(p-1)$ sub-matrix consisting of the remaining columns.

\begin{proposition}
\label{propositionNB1}
Let Assumptions~\ref{assumptionNB1} and~\ref{assumptionNB2} be satisfied, we have
\[
\max_{1 \leq j \leq p}\left\|\widehat{\theta}_{j}-\theta_{j}^{*}\right\|_{\infty} \leq \tilde c_2  \sqrt{\frac{\log p}{n}} \; , 
\]
\end{proposition}
with probability at least $1-  p ^{2-A^{2} / 8}$ where  $\tilde c_2 = A \sigma \cdot \frac{3}{2}\left(1+\frac{\left(1+c_{0}\right)^{2}}{\left(1+2 c_{0}\right)(\alpha-1)}\right)$ and $A>4$.

\noindent \underline{\textbf{Proof of Proposition~\ref{propositionNB1}:}}
By Theorem 1 of \citet{lounici2008sup}, we have:
$$
\mathbb{P}\left( \left\|\widehat{\theta}_{j}-\theta_j^{*}\right\|_{\infty} \leq c_{2} r\right) \geq 1 - p^{1-A^{2} / 8} , 
$$
with  $r=A \sigma \sqrt{(\log p) / n}$ and  $c_{2}=\frac{3}{2}\left(1+\frac{\left(1+c_{0}\right)^{2}}{\left(1+2 c_{0}\right)(\alpha-1)}\right)$. Here, we approximate $p-1$ by $p$ since $p$ is sufficiently large.

Or equivalently, 
$$
\mathbb{P}\left( \left\|\widehat{\theta}_{j}-\theta_j^{*}\right\|_{\infty} \geq c_{2} r\right) \leq  p^{1-A^{2} / 8}.
$$
By union bound, we have:
$$
\mathbb{P}\left( \max \limits_{ 1 \leq j \leq p} \left\|\widehat{\theta}_{j}-\theta_j^{*}\right\|_{\infty} \geq c_{2} r\right) \leq  p \cdot  p^{1-A^{2} / 8}  = p ^{2-A^{2} / 8} .
$$
Hence, if we choose $A > 4$, we have:
$$
\mathbb{P}\left( \max \limits_{ 1 \leq j \leq p} \left\|\widehat{\theta}_{j}-\theta_j^{*}\right\|_{\infty} \leq c_{2} r\right) \geq  1-  p ^{2-A^{2} / 8} \to 1.
$$
Or
$$
\mathbb{P}\left( \max \limits_{ 1 \leq j \leq p} \left\|\widehat{\theta}_{j}-\theta_j^{*}\right\|_{\infty} \leq \tilde c_{2} \sqrt{\frac{\log p}{n}} \right) \geq  1-  p ^{2-A^{2} / 8} \to 1,
$$
where $\tilde c_2 = A \sigma \cdot \frac{3}{2}\left(1+\frac{\left(1+c_{0}\right)^{2}}{\left(1+2 c_{0}\right)(\alpha-1)}\right)$. $\qed$

To establish graphical model selection consistency, we further assume the minimum signal strength condition.

\begin{assumption}
\label{assumptionNB3}
Define the minimum signal strength:
 \begin{equation*}
 \theta_{\min }:= \min _{1 \leq j \leq p} \min _{j, k \in N(j)}\left|\left(\theta_{j}^{*}\right)_{k}\right|  > c_1 \sqrt{\frac{\log p}{n}} \;\;.   
\end{equation*}
\end{assumption}

\begin{proposition}
\label{propositionNB2}
Let Assumptions~\ref{assumptionNB1} and~\ref{assumptionNB2} be satisfied. Let Assumption~\ref{assumptionNB3} be satisfied with $c_1 > 2 \tilde c_2$.  The thresholded neighborhood selection estimate $\widetilde{\bTheta}^{\lambda,\tau}$ with threshold level $\tau = \tilde c_2 \sqrt{\frac{\log p}{n}}$ satisfies:
\[
 \operatorname{sign} ( \tilde \theta_{j} ) = \operatorname{sign} ( \theta_{j}^{*} ), \hspace{6mm} \forall 1 \leq  j \leq p,
 \]
with probability at least $1-  p ^{2-A^{2} / 8}$ where  $\tilde c_2 = A \sigma \cdot \frac{3}{2}\left(1+\frac{\left(1+c_{0}\right)^{2}}{\left(1+2 c_{0}\right)(\alpha-1)}\right)$ and $A>4$.
\end{proposition}

Note \citet{meinshausen2009lasso} also proved model selection consistency of thresholded Lasso estimator by establishing $\ell_2$-norm convergence rate which requires incoherent design associated with eigenvalues of the covariance matrix. In particular, they show that, under the incoherent design condition with a sparsity multiplier sequence $e_{n}$, if $\lambda \propto \sigma e_{n} \sqrt{\log p_{n}/n}$, there exists a constant $M>0$ such that
$$
\left\|\hat{\beta}^{\lambda_{n} }- \beta^*  \right\|_2^2 \leq M \sigma^{2} \frac{s_{n} \log p_{n}}{n} \frac{e_{n}^{2}}{\phi_{\min }^{2}\left(e_{n}^{2} s_{n}\right)} \;\;,
$$
with probability converging to 1 for $n \rightarrow \infty$.
We can follow the similar approach we use in proving Theorem~\ref{theorem1} and show graphical model selection consistency under incoherent design assumption.

\section{Graphical Lasso In the Presence of Latent Variables}\label{appen_glasso_lv}
 
In this section, we study the graphical model selection consistency of graphical Lasso in the presence of latent variables.  In addition to  Assumptions~\ref{assumptionB1}-\ref{assumptionB3}  required for graphical Lasso to establish graphical model selection consistency without latent variables suggested by \citet{ravikumar2011high}, we have Assumption~\ref{assumptionB4}, which requires that the quantity associated with the latent variables, has to vanish.
\begin{assumption}
\label{assumptionB1}
Suppose that the variables $X_i/\sqrt{\Sigma^*_{ii}}$ are sub-Gaussian with parameter $\sigma$, where $\Sigma^* = (S^* - L^*)^{-1}$.
\end{assumption}
\begin{assumption}
\label{assumptionB2} (Irrepresentable condition)
    Denote:
    $\Gamma^{*} :=\left.\nabla_{\Theta}^{2} g(\Theta)\right|_{\Theta=S^{*}}   = (S^*)^{-1}    \otimes  (S^*)^{-1}  =  \left( S^* \otimes S^* \right)^{-1}$. There exists some $\alpha \in(0,1]$ such that
\begin{equation*}
    \max_{e \in S^{c}}\left\|\Gamma_{e S}^{*}\left(\Gamma_{S S}^{*}\right)^{-1}\right\|_{1} \leq(1-\alpha).
\end{equation*}
\end{assumption}
\begin{assumption}
\label{assumptionB3}
Define the minimum signal strength:
\begin{equation*}
\theta_{\min }:=\min _{(i, j) \in E\left(S^{*}\right)} \left|S_{i j}^{*}\right|>     \left\{16 \sqrt{2}\left(1+4 \sigma^{2}\right) \max _{i}\left(\Sigma_{i i}^{*}\right)\left(1+12 \alpha^{-1}+ C_2\right) \kappa_{\Gamma^{*}}  \right\} \sqrt{\frac{\tau \log p+\log 4}{n}}   .       
\end{equation*}

\end{assumption}

\hide{
\begin{assumption}
\label{assumptionB4}
$\|  (S^* - L^*)^{-1} -  (S^*)^{-1}   \|_{\infty}  = \mathcal O\bigg( \sqrt{\frac{\log p}{n}} \bigg)$.
\end{assumption}
}
\assumptionBfour*

 \hide{
\begin{theorem}
\label{theorem2}
Let Assumptions~\ref{assumptionB1}-\ref{assumptionB3} and Assumption~\ref{assumptionB4} be satisfied.
Then, if the sample size $n$ satisfies the bound
\begin{align}
n > C_{1} d^{2}\left(1+\frac{12}{\alpha}+C_2\right)^{2}(\tau \log p+\log 4) \label{eq:31},
\end{align}
then with probability greater than $1 - 1/p^{\tau - 2}$, 
the graphical Lasso estimator $\widehat \Theta^{\lambda}$ with regularization parameter $\lambda = (12/\alpha) \bar{\delta}_{f}\left(n, p^{\tau}\right)$ is model selection consistent with high probability as $p \to \infty$,
\begin{align*}
\mathbb{P}\left(\operatorname{sign}(\widehat{\Theta}^{\lambda}_{ij})=\operatorname{sign}\left(S_{ij}^{*}\right), \forall \widehat{\Theta}^{\lambda}_{ij} \in \widehat{\Theta}^{\lambda}\right) \geq 1-1 / p^{\tau-2} \rightarrow 1. 
\end{align*}
\end{theorem}
}
\theoremtwo*

Here, $C_1 = \left\{48 \sqrt{2}\left(1+4 \sigma^{2}\right) \max _{i}\left(\Sigma_{i i}^{*}\right) \max \left\{\kappa_{{S^*}^{-1}} \kappa_{\Gamma^{*}}, \kappa_{{S^*}^{-1}}^{3} \kappa_{\Gamma^{*}}^{2}\right\}\right\}^{2}$, $C_2 = \|  (S^* - L^*)^{-1} -  (S^*)^{-1}   \|_{\infty}  / (\frac{\alpha \lambda}{12})$ and $\bar{\delta}_{f}\left(n, p^{\tau}\right) = \sqrt{128\left(1+4 \sigma^{2}\right)^{2} \max _{i}\left(\Sigma_{i i}^{*}\right)^{2}} \sqrt{\frac{\tau \log p+\log 4}{n}}$.
 
We follow closely the proof approach used by \citet{ravikumar2011high}. In particular, we also use primal-dual witness approach. To begin with, recall the graphical Lasso problem is:
$$
\widehat \Theta :=\argmin _{\Theta \in \mathcal S_{++}^p } \left\{\langle\langle\Theta, \widehat{\Sigma}\rangle\rangle-\log \operatorname{det}(\Theta)+\lambda_{n}\|\Theta\|_{1,\text {off}}\right\},
$$
where $\langle\langle U,V \rangle\rangle := \sum_{i,j} U_{ij} V_{ij}$ is the trace inner product on the space of symmeric matrices.

We denote the sub-differential of the norm $\| \cdot \|_{1,\text{off}}$:
\begin{align*}
Z_{i j}=
\begin{cases}
0      & \text { if } i=j \\
\operatorname{sign}\left(\Theta_{i j}\right) &  \text { if } i \neq j \text { and } \Theta_{i j} \neq 0 \\
\in[-1,+1] & \text { if } i \neq j \text { and } \Theta_{i j}=0. \\
\end{cases}
\end{align*}

First, we have the optimality condition of the graphical Lasso problem. 

\begin{lemma}[\citealt{ravikumar2011high}, Lemma 3]
\label{lemma3_rav}
For any $\lambda_{n}>0$ and sample covariance $\widehat \Sigma$ with strictly positive diagonal elements, the $\ell_1$-regularized log-determinant problem has a unique solution $\widehat{\Theta} \succ 0$ characterized by
\begin{align}
\widehat{\Sigma}-\widehat{\Theta}^{-1}+\lambda_{n} \widehat{Z} = 0,  \label{eq:46}
\end{align}
where $\widehat{Z}$ is an element of the subdifferential $\partial\|\widehat{\Theta}\|_{1,\text{off}}$.   
\end{lemma}
Lemma~\ref{lemma3_rav} is the same as  Lemma 3 by \citet{ravikumar2011high}. We first introduce some notations before adopting the primal-dual witness approach. Recall the true edge set is denoted as $E\left(\Theta^{*}\right)=\left\{ (i, j) \in V \times V \mid i \neq j, \Theta_{i j}^{*} \neq 0\right\}$. We denote $S\left(\Theta^{*}\right)= E\left(\Theta^{*}\right) \cup\{(1,1), \ldots,(p, p) \}  $ as the set including the diagonal elements. We use $S\left(\Theta^{*}\right)^{c}$ to denote the complement of $S\left(\Theta^{*}\right)$; we use $S$ and $S^{c}$ for simplicity respectively. Here, with a slight abuse of notation, $S^*$ refers to the true sparse matrix while subscript $S$ refers to the active set.

Based on this lemma, we construct the primal-dual witness solution $(\widetilde \Theta, \widetilde Z)$ like \citet{ravikumar2011high} as follows:

(a) We determine the matrix $\widetilde{\Theta}$ by solving the restricted log-determinant problem
$$
\widetilde{\Theta}:=\argmin _{\Theta \succ 0, \Theta=\Theta^{T}, \Theta_{S^{c}=0}} \left\{\langle\langle\Theta, \widehat{\Sigma}\rangle\rangle-\log \operatorname{det}(\Theta)+\lambda_{n}\|\Theta\|_{1,\text {off}}\right\}.
$$
Note that by construction, we have $\widetilde{\Theta} \succ 0,$ and moreover $\widetilde{\Theta}_{S^{c}}=0$.

(b) We choose $\widetilde{Z}$ as a member of the sub-differential of the regularizer $\|\cdot\|_{1,\text{off}}$ evaluated at $\widetilde{\Theta}$.

(c) For each $(i, j) \in S^{c},$ we replace $\widetilde{Z}_{i j}$ with the quantity
$$
\widetilde{Z}_{i j}:=\frac{1}{\lambda_{n}}\left\{-\widehat{\Sigma}_{i j}+\left[\widetilde{\Theta}^{-1}\right]_{i j}\right\},
$$
which ensures that constructed matrices $(\widetilde{\Theta}, \widetilde{Z})$ satisfy the optimality condition~\eqref{eq:46}.

(d) We verify the strict dual feasibility condition
$$
\left|\widetilde{Z}_{ij}\right|<1 \quad \text { for all }(i, j) \in S^{c}.
$$

Slightly different from the notation by \citet{ravikumar2011high}, we denote $W$ as the ``effective noise" in the sample covariance matrix $\Sigma$:
$$
W:=\widehat{\Sigma}-\left(S^{*} - L^{*}\right)^{-1}.
$$
Next, we use $\Delta=\widetilde{\Theta}-S^{*}$  to measure the difference between the primal witness matrix $\widetilde{\Theta}$ and the ground truth $S^{*}$. Note that by the definition of $\widetilde{\Theta}$, we have $\Delta_{S^{c}}=0$. 
Finally, we denote $R(\Delta)$ as the difference of the gradient $\nabla g(\widetilde{\Theta})=\widetilde{\Theta}^{-1}$ from its first-order Taylor expansion around $S^*$:
$$
R(\Delta)=\widetilde{\Theta}^{-1}-S^{*-1}+S^{*-1} \Delta S^{*-1}.
$$
We have the following Lemma~\ref{lemma4_rav}, providing sufficient condition for strict dual feasibility to hold, i.e., $\left\|\widetilde{Z}_{S^{c}}\right\|_{\infty}<1$.
\begin{lemma}[Strict dual feasibility]
\label{lemma4_rav}
Suppose that
\begin{align}
\max \left\{\|W\|_{\infty},\|R(\Delta)\|_{\infty}, \| \left(S^{*} - L^{*}\right)^{-1} - {S^{*}}^{-1} \|_{\infty} \right\} \leq \frac{\alpha \lambda_{n}}{12}. \label{eq:51}
\end{align}
Then the vector $\widetilde{Z}_{S^{c}}$ constructed in step (c) 
satisfies $\left\|\widetilde{Z}_{S^{c}}\right\|_{\infty}<1$, and therefore
$\widetilde{\Theta}=\widehat{\Theta}$.
\end{lemma}

\noindent \underline{\textbf{Proof of Lemma~\ref{lemma4_rav}:}}
We follow the proof technique of Lemma 4 of \citet{ravikumar2011high}; we include the term associated with the latent variables,  $ \| \left(S^{*} - L^{*}\right)^{-1} - {S^{*}}^{-1} \|_{\infty}$ in our proof.

By the construction of $W$ and $R(\Delta)$, we can re-write the stationary condition, $\widehat{\Sigma}-\widetilde{\Theta}^{-1}+\lambda_{n} \widetilde{Z}=0$ equivalently as:
\begin{align*}
    &W + \left(S^{*} - L^{*}\right)^{-1}-\widetilde{\Theta}^{-1}+\lambda_{n} \widetilde{Z}=0.
\end{align*}
Denote $\left(S^{*} - L^{*}\right)^{-1} - {S^{*}}^{-1}$ as $Q^{*}$, we have
\begin{align}
    & W + {S^{*}}^{-1} + Q^{*} -\widetilde{\Theta}^{-1}+\lambda_{n} \widetilde{Z}=0   \nonumber \\
    & W + Q^{*}    + S^{*-1} \Delta S^{*-1} - R(\Delta)      +\lambda_{n} \widetilde{Z}=0.  \label{eq:52}
\end{align}

We can re-write the above matrix equality  as an ordinary linear equation by vectorizing the matrices.  We use the notation $\operatorname{vec}(A)$ or equivalently $\overline{A}$ for the vector version of the set or matrix $A$ obtained by concatenating the rows of $A$ into a single column vector.
$$
\operatorname{vec}\left(S^{*-1} \Delta S^{*-1}\right)=\left(S^{*-1} \otimes S^{*-1}\right) \bar{\Delta}=\Gamma^{*} \bar{\Delta}.
$$
By the disjoint decomposition $S$ and $S^{c}$, equation~\eqref{eq:52} can be re-written as two blocks of linear equations as follows:
\begin{align}
& \Gamma_{S S}^{*} \overline{\Delta}_{S}+\overline{W}_{S} + \overline Q_{S}^{*}  -\overline{R}_{S}+\lambda_{n} \overline{\widetilde Z}_{S} =0  \label{eq:53a} \\
& \Gamma_{S^{c} S}^{*} \overline{\Delta}_{S}+ \overline W_{S^{c}} + \overline Q_{S^{c}}^{*} - \overline R_{S^{c}}+\lambda_{n} \overline{\widetilde Z}_{S^{c}} =0.  \label{eq:53b}
\end{align}

Here we use the fact that $\Delta_{S^{c}}=0$ by construction. Since $\Gamma_{SS}^{*}$ is invertible, we can solve for $\overline{\Delta}_{S}$ from equation~\eqref{eq:53a} as follows:
$$
\overline{\Delta}_{S}=\left(\Gamma_{S S}^{*}\right)^{-1}\left[-\overline{W}_{S} -  \overline Q_{S}^{*}  +\overline{R}_{S}-\lambda_{n} \overline{\widetilde Z}_{S}\right].
$$

Substituting this expression into equation~\eqref{eq:53b}, we can solve for $\widetilde{Z}_{S^{c}}$ as follows:
$$
\begin{aligned}
\overline{\widetilde{Z}}_{S^{c}}=&-\frac{1}{\lambda_{n}} \Gamma_{S^{c} S}^{*} \overline{\Delta}_{S}+\frac{1}{\lambda_{n}} \overline{R}_{S^{c}}-\frac{1}{\lambda_{n}} \overline{W}_{S^{c}} - \frac{1}{\lambda_{n}} \overline{Q}^*_{S^{c}}  \\
=&-\frac{1}{\lambda_{n}} \Gamma_{S^{c} S}^{*}\left(\Gamma_{S S}^{*}\right)^{-1}\left(\overline W_{S}+\overline{Q}^*_{S} - \overline R_{S}\right)+\Gamma_{S^{c} S}^{*}\left(\Gamma_{S S}^{*}\right)^{-1} \overline{\widetilde Z}_{S} \\
&-\frac{1}{\lambda_{n}}\left(\overline{W}_{S^{c}}+\overline{Q}^*_{S^{c}} - \overline{R}_{S^{c}}\right).
\end{aligned}
$$

Taking the $\ell_{\infty}$ norm of both sides yields
$$
\begin{aligned}
\left\|\overline{\widetilde Z}_{S^{c}}\right\|_{\infty} &\leq \frac{1}{\lambda_{n}}\vertiii{\Gamma_{S^{c} S}^{*}\left(\Gamma_{S S}^{*}\right)^{-1}}_{\infty}\left(\left\|\overline{W}_{S}\right\|_{\infty}+    \left\|\overline{Q}^*_{S}\right\|_{\infty} +  \left\|\overline{R}_{S}\right\|_{\infty}\right)  \\
&+\left\|\Gamma_{S^{c} S}^{*}\left(\Gamma_{S S}^{*}\right)^{-1} \overline{\widetilde{Z}}_{S}\right\|_{\infty}+\frac{1}{\lambda_{n}}\left(\left\|\overline{W}_{S^{c}}\right\|_{\infty}+ \left\|\overline{Q}^*_{S^{c}}\right\|_{\infty}  + \left\|\overline{R}_{S^{c}}\right\|_{\infty}\right).
\end{aligned}
$$

Since $\widetilde{Z}$ belongs to the subdifferential of the norm $\|\cdot\|_{1, \text { off }}$ by construction, we have $\left\|\overline{\widetilde Z}_{S}\right\|_{\infty} \leq 1$.  By Assumption~\ref{assumptionB2}, we have that $\left\|\Gamma_{S^{c} S}^{*}\left(\Gamma_{S S}^{*}\right)^{-1} \overline{\widetilde{Z}}_{S^{c}}\right\|_{\infty} \leq(1-\alpha)$. Hence,
we have
$$
\left\|\overline{\widetilde Z}_{S c}\right\|_{\infty} \leq \frac{2-\alpha}{\lambda_{n}}\left(\left\|\overline{W} \right\|_{\infty}+ \left\|\overline{Q}^* \right\|_{\infty}  + \left\|\overline{R} \right\|_{\infty}\right)+(1-\alpha).
$$

Finally, applying the assumption in Lemma~\ref{lemma4_rav}, we have
$$
\left\|\overline{\widetilde{Z}}_{S^{c}}\right\|_{\infty} \leq \frac{(2-\alpha)}{\lambda_{n}}\left(\frac{\alpha \lambda_{n}}{4}\right)+(1-\alpha) \leq \frac{\alpha}{2}+(1-\alpha)<1,
$$
as claimed.  $\qed$

Next, we relate the behavior of the remainder term $R(\Delta)$ to the deviation
$ \Delta=\widetilde{\Theta}-S^{*}$.

\begin{lemma}[Control of remainder]
\label{lemma5_rav}
Suppose that the elementwise $\ell_{\infty}$-bound $\|\Delta\|_{\infty} \leq \frac{1}{3 \kappa_{ {S^*}^{-1} }d}$ holds. Then the matrix $J:=\sum_{k=0}^{\infty}(-1)^{k}\left({S^*}^{-1} \Delta\right)^{k}$ satisfies
the $\ell_{\infty}$-operator norm $\left\|J^{T}\right\|_{\infty} \leq 3 / 2,$ and moreover, the matrix
\begin{align}
R(\Delta)={S^*}^{-1} \Delta {S^*}^{-1} \Delta J {S^*}^{-1} \label{eq:56}
\end{align}
has elementwise $\ell_{\infty}$-norm bounded as
\begin{align}
\|R(\Delta)\|_{\infty} \leq \frac{3}{2} d\|\Delta\|_{\infty}^{2}  \kappa_{{S^{*}}^{-1}}^{3}, \label{eq:57}
\end{align}
where $\kappa_{{S^*}^{-1}}:=\vertiii{S^*}_{\infty}=\left(\max _{i=1, \ldots, p} \sum_{j=1}^{p}\left|({S^*}^{-1})_{i j}\right|\right)$.
\end{lemma}

\noindent \underline{\textbf{Proof of Lemma~\ref{lemma5_rav}:}}
We follow the proof technique of Lemma 5 of \citet{ravikumar2011high}.

We rewrite the remainder as:
\begin{align*}
    R(\Delta)&=\widetilde{\Theta}^{-1}-S^{*-1}+S^{*-1} \Delta S^{*-1} \\
     &= (S^* + \Delta)^{-1}-S^{*-1}+S^{*-1} \Delta S^{*-1}.
\end{align*}

\hide{
Note that:
\begin{align*}
\left\|S^{*-1} \Delta\right\|_{\infty} & \leq\left\|S^{*-1}\right\|_{\infty}\|\Delta\|_{\infty} \\
& \leq \kappa_{{S^*}^{-1}} d\|\Delta\|_{\infty}<1 / 3,
\end{align*}
}

By sub-multiplicativity of the $\vertiii{\cdot}_{\infty}$ matrix norm, for any two matrices $A$ and $B$, we have $\vertiii{AB}_{\infty} \leq \vertiii{A}_{\infty} \vertiii{B}_{\infty}$, so that
\begin{align*}
\vertiii{S^{*-1} \Delta}_{\infty} & \leq \vertiii{S^{*-1}}_{\infty}  \vertiii{\Delta}_{\infty} \\
& \leq \kappa_{{S^*}^{-1}} d\|\Delta\|_{\infty}<1 / 3,
\end{align*}
where we use the definition of $\kappa_{{S^*}^{-1}}$, the fact that $\Delta$ has at most $d$ non-zeros per row/column, and the assumption $\|\Delta\|_{\infty} \leq \frac{1}{3 \kappa_{ {S^*}^{-1} }d}$.

The rest of the proof follows the proof of Lemma 5 of \citet{ravikumar2011high} using matrix algebra. We have:
\begin{align*}
\|R(\Delta)\|_{\infty} & \leq \frac{3}{2}\|\Delta\|_{\infty} \kappa_{{S^*}^{-1}}\left\|S^{*-1}\right\|_{\infty}^{2}\|\Delta\|_{\infty} \\
& \leq \frac{3}{2} d\|\Delta\|_{\infty}^{2} \kappa_{{S^*}^{-1}}^{3}.
\end{align*}
$\qed$

Next, we state Lemma~\ref{lemma6_rav}, which gives the $\ell_{\infty}$-norm bound on the deviation $\widetilde{\Theta}-S^{*}$.

\begin{lemma}[Control of $\Delta$]
\label{lemma6_rav}
Suppose that
\begin{align}
r:=2 \kappa_{\Gamma^{*}}\left(\|W\|_{\infty}+\lambda_{n} +    \|  (S^* - L^*)^{-1} -  (S^*)^{-1}   \|_{\infty}     \right) \leq \min \left\{\frac{1}{3 \kappa_{{S^{*}}^{-1}} d} \frac{1}{3 \kappa_{{S^{*}}^{-1}}^{3} \kappa_{\Gamma^{*}} d}\right\}.  \label{eq:58}
\end{align}
Then we have the elementwise $\ell_{\infty}$ bound
$$
\|\Delta\|_{\infty}=\left\|\widetilde{\Theta}-S^{*}\right\|_{\infty} \leq r.
$$
\end{lemma}

\noindent \underline{\textbf{Proof of Lemma~\ref{lemma6_rav}:}}
We follow the proof technique of Lemma 6 of \citet{ravikumar2011high}; we include the term associated with the latent variables,  $ \| \left(S^{*} - L^{*}\right)^{-1} - {S^{*}}^{-1} \|_{\infty}$ in our proof.

If we take partial derivatives of the Lagrangian of the restricted problem with respect to the unconstrained elements $\Theta_S$, we have the zero-gradient condition:
\begin{align}
G\left(\Theta_{S}\right)=-\left[\Theta^{-1}\right]_{S}+\widehat{\Sigma}_{S}+\lambda_{n} \tilde{Z}_{S}=0.   \label{eq:70}
\end{align}

Our goal is to bound the deviation $\Delta=\widetilde{\Theta}-S^{*}$.  The strategy is to show the existence of a solution $\Delta$ to the zero-gradient condition~\eqref{eq:70} that is contained inside the ball $\mathbb{B}(r)$ defined as:
\begin{align}
\mathbb{B}(r):=\left\{\Theta_{S} \mid\left\|\Theta_{S}\right\|_{\infty} \leq r\right\}, \quad \text { with } r:=2 \kappa_{\Gamma^{*}}\left(\|W\|_{\infty}+\lambda_{n} +    \|  (S^* - L^*)^{-1} -  (S^*)^{-1}   \|_{\infty}  \right).   \label{eq:60}
\end{align}

By uniqueness of the optimal solution, we can thus conclude that $\widetilde{\Theta}-S^{*}$ belongs to this ball. In terms of the vector $\bar \Delta_S = \overline{\tilde \Theta_S} - \overline{S^*_S}$,  we define a map $F$ via:
$$
F\left(\overline{\Delta}_{S}\right):=-\left(\Gamma_{S S}^{*}\right)^{-1}\left(\overline{G}\left(S_{S}^{*}+\Delta_{S}\right)\right)+\overline{\Delta}_{S},
$$
where $\overline G$ refers to the vectorized version of $G$. Note that by construction $F(\overline \Delta_S) = \overline \Delta_S$ holds if and only if $G(S_S^* + \Delta_S) = G(\tilde \Theta_S) = 0$. Recall that, with a slight abuse of notation, $S^*$ refers to the true sparse matrix while subscript $S$ refers to the active set.

Next, we show $F(\mathbb B(r) \subseteq \mathbb B(r)$. Since $F$ is continuous and $\mathbb B(r)$ is convex and compact. By Brouwer's fixed point theorem \citep{ortega2000iterative}, there exists some fixed point $\overline \Delta_S \in \mathbb{B}(r)$. By uniqueness of the zero gradient condition, we conclude that $\|\tilde \Theta_S - S_S^* \|_{\infty} \leq r$.

By definition, we have:
\begin{align*}
G\left(S_{S}^{*}+\Delta_{S}\right) &=-\left[\left(S^{*}+\Delta\right)^{-1}\right]_{S}+\widehat{\Sigma}_{S}+\lambda_{n} \widetilde{Z}_{S} \\
&=\left[-\left[\left(S^{*}+\Delta\right)^{-1}\right]_{S}+\left[ {S^{*}}^{-1}\right]_{S}\right]+\left[\widehat{\Sigma}_{S}-\left[ {S^{*}}^{-1}\right]_{S}\right]+\lambda_{n} \widetilde{Z}_{S} \\
&=\left[-\left[\left(S^{*}+\Delta\right)^{-1}\right]_{S}+\left[ {S^{*}}^{-1}\right]_{S}\right]+W_{S}+\lambda_{n} \tilde{Z}_{S} +  \left[\left(S^*-L^*\right)^{-1} \right]_S -  \left[ {S^{*}}^{-1}\right]_S,
\end{align*}
where we use the fact $W =\widehat{\Sigma}-\left(S^{*} - L^{*}\right)^{-1}$.

By definition~\eqref{eq:60} of the radius $r$ and the assumed upper bound\eqref{eq:58}, we have $\| \Delta \|_{\infty} \leq r \leq \frac{1}{3 \kappa_{ {S^*}^{-1} } d}$. Therefore, the results of Lemma~\ref{lemma5_rav} apply. Using the definition of the remainder, taking the vectorized form of expression~\eqref{eq:56} and restricting to entries in $S$, we have:
$$
\operatorname{vec}\left(\left(S^{*}+\Delta\right)^{-1}-S^{*-1}\right)_{S}+\Gamma_{S S}^{*} \overline{\Delta}_{S}=\operatorname{vec}\left(\left(S^{*-1} \Delta\right)^{2} J S^{*-1}\right)_{S}.
$$
Combine this with the expression for $G$, we have:
\begin{align*}
F\left(\overline{\Delta}_{S}\right) &=-\left(\Gamma_{S S}^{*}\right)^{-1} \bar{G}\left(S_{S}^{*}+\Delta_{S}\right)+\bar{\Delta}_{S} \\
&=\left(\Gamma_{S S}^{*}\right)^{-1} \operatorname{vec}\left\{\left[\left(S^{*}+\Delta\right)^{-1}-S^{*-1}\right]_{S}-W_{S}-\lambda_{n} \widetilde{Z}_{S}\right\}+\overline{\Delta}_{S} \\
&=\underbrace{\left(\Gamma_{S S}^{*}\right)^{-1} \operatorname{vec}\left[\left(S^{*-1} \Delta\right)^{2} J S^{*-1}\right]_{S}}_{T_1}   -     \underbrace{\left(\Gamma_{S S}^{*}\right)^{-1}\left(\overline{W}_{S}+\lambda_{n} \overline{\widetilde{Z}}_{S} +  \overline{\left[\left(S^*-L^*\right)^{-1} \right]}_S -  \overline{\left[ {S^{*}}^{-1}\right]}_S \right)}_{T_2}.
\end{align*}
For the second term, by the definition of $\kappa_{\Gamma^{*}} = \vertiii{(\Gamma_{S S}^{*})^{-1}}_{\infty}$, we have $$\|T_2 \|_{\infty} \leq \kappa_{\Gamma^{*}}\left(\|W\|_{\infty}+\lambda_{n} +    \|  (S^* - L^*)^{-1} -  (S^*)^{-1}   \|_{\infty}     \right) = r/2.$$ It suffices to show that $\|T_1\|_{\infty} \leq r/2$. We have
$$
\begin{aligned}
\left\|T_{1}\right\|_{\infty} & \leq \kappa_{\Gamma^{*}}\left\|\operatorname{vec}\left[\left({S^{*}}^{-1} \Delta\right)^{2} J {S^{*}}^{-1}\right]_{S}\right\|_{\infty} \\
& \leq \kappa_{\Gamma^{*}}\|R(\Delta)\|_{\infty},
\end{aligned}
$$
where we used the expanded form~\eqref{eq:56} of the remainder. Applying the bound~\eqref{eq:57} from Lemma~\ref{lemma5_rav}, we obtain
$$
\left\|T_{1}\right\|_{\infty} \leq \frac{3}{2} d \kappa_{{S^*}^{-1}}^{3} \kappa_{\Gamma^{*}}\|\Delta\|_{\infty}^{2} \leq \frac{3}{2} d \kappa_{{S^*}^{-1}}^{3} \kappa_{\Gamma^{*}} r^{2}.
$$
Since $r \leq \frac{1}{3 \kappa_{{S^*}^{-1}}^{3}\kappa_{\Gamma^{*}} d}$ by assumption~\eqref{eq:58}, we conclude that
$$
\left\|T_{1}\right\|_{\infty} \leq \frac{3}{2} d \kappa_{{S^*}^{-1}}^{3} \kappa_{\Gamma^{*}} \frac{1}{3 \kappa_{{S^*}^{-1}}^{3}\kappa_{\Gamma^{*}} d} r=r / 2,
$$
thereby establishing the claim.
$\qed$


We control the sampling noise $W =\widehat{\Sigma} - \Sigma^* = \widehat{\Sigma}-\left(S^{*} - L^{*}\right)^{-1}$. This control is specified in terms of the decay function $f$.

\begin{lemma}[Control of Sampling Noise, \citealt{ravikumar2011high}, Lemma 8]
\label{lemma8_rav}
For any $\tau>2$ and sample size $n$ such that $\bar{\delta}_{f}\left(n, p^{\tau}\right) \leq 1 / v_{*},$ we have
$$
\mathbb{P}\left[\|W\|_{\infty} \geq \bar{\delta}_{f}\left(n, p^{\tau}\right)\right] \leq \frac{1}{p^{\tau-2}} \rightarrow 0.
$$
\end{lemma}

Lemma~\ref{lemma8_rav} is the same as  Lemma 8 by \citet{ravikumar2011high}.

Note by Lemma 1 of \citet{ravikumar2011high}, entries of the sample covariance based on i.i.d. samples of sub-Gaussian random vector satisfy an exponential-type tail bound. Here,  $X_i/\sqrt{\Sigma^*_{ii}}$ are sub-Gaussian with parameter $\sigma$. Hence, similar to  Corollary 1 of \citet{ravikumar2011high}, the inverse function $\bar{\delta}_{f}\left(n, p^{\tau}\right)$ takes the form:
$$\bar{\delta}_{f}\left(n, p^{\tau}\right) =\sqrt{128\left(1+4 \sigma^{2}\right)^{2} \max _{i}\left(\Sigma_{i i}^{*}\right)^{2}} \sqrt{\frac{\tau \log p+\log 4}{n}}.$$

\noindent \underline{\textbf{Proof of Theorem~\ref{theorem2}:}}
We follow the proof technique of Theorem 1 of \citet{ravikumar2011high}. We first show that with high probability the witness matrix $\widetilde{\Theta}$ is equal to the solution $\widehat{\Theta}$ to the original log-determinant problem.

Let $\mathcal{A}$ denote the event that $\|W\|_{\infty} \leq \bar{\delta}_{f}\left(n, p^{\tau}\right)$. Using the monotonicity of the inverse tail function, the lower lower bound on the sample size $n$ implies that $\bar{\delta}_{f}\left(n, p^{\tau}\right) \leq 1 / v_{*}$. Consequently, Lemma~\ref{lemma8_rav} implies that $\mathbb{P}(\mathcal{A}) \geq 1-\frac{1}{p^{\tau-2}}$.

Next we verify the assumption~\eqref{eq:51} of Lemma~\ref{lemma4_rav} holds. Recall the choice of regularization penalty $\lambda_{n}=(12 / \alpha) \bar{\delta}_{f}\left(n, p^{\tau}\right),$ we have $\|W\|_{\infty} \leq$ $(\alpha / 12) \lambda_{n} .$ In order to establish condition~\eqref{eq:51} it remains to establish the bound $\|R(\Delta)\|_{\infty} \leq \frac{\alpha \lambda_{n}}{12} .$ We do so in two steps, by using Lemmas~\ref{lemma6_rav} and~\ref{lemma5_rav} consecutively. First, we show that the condition~\eqref{eq:58} required for Lemma~\ref{lemma6_rav} to hold is satisfied under the specified conditions on $n$ and $\lambda_{n}$. Also, by Assumption~\ref{assumptionB4}, $\|  (S^* - L^*)^{-1} -  (S^*)^{-1}   \|_{\infty}  = \mathcal O\bigg( \sqrt{\frac{\log p}{n}} \bigg)$. Therefore, by the construction of $\bar{\delta}_{f}$, there exists a constant $C_2$ such that $\|  (S^* - L^*)^{-1} -  (S^*)^{-1}   \|_{\infty}  = C_2 \bar{\delta}_{f}\left(n, p^{\tau}\right)$.

From Lemma~\ref{lemma8_rav} and our choice of regularization constant $\lambda_{n}=(12 / \alpha) \bar{\delta}_{f}\left(n, p^{\tau}\right)$,
$$
2 \kappa_{\Gamma^{*}}\left(\|W\|_{\infty}+\lambda_{n} +   \|  (S^* - L^*)^{-1} -  (S^*)^{-1}   \|_{\infty} \right) \leq 2 \kappa_{\Gamma^{*}}\left(1+\frac{12}{\alpha}+C_2\right) \bar{\delta}_{f}\left(n, p^{\tau}\right).
$$

The lower bound~\eqref{eq:31} is equivalent to 
\begin{align}
    n>\bar{n}_{f} \bigg(1 / \max \Big\{v_{*}, 6\left(1+12 \alpha^{-1}+C_2 \right) d \max \left\{\kappa_{{S^*}^{-1}}  \kappa_{\Gamma^{*}}, \kappa_{{S^*}^{-1}} ^{3} \kappa_{\Gamma^{*}}^{2}\right\}\Big\}, p^{\tau}\bigg), \label{eq:29}
\end{align}
as suggested by \citet{ravikumar2011high}.

From the lower bound~\eqref{eq:29} and the monotonicity of the tail inverse functions, we have
\begin{align}
2 \kappa_{\Gamma^{*}}\left(1+\frac{12}{\alpha}+C_2\right) \bar{\delta}_{f}\left(n, p^{\tau}\right) \leq \min \left\{\frac{1}{3 \kappa_{{S^*}^{-1}} d}, \frac{1}{3 \kappa_{{S^*}^{-1}}^{3} \kappa_{\Gamma^{*}} d}\right\},  \label{eq:63}
\end{align}
showing that the assumptions of Lemma~\ref{lemma6_rav} are satisfied. Applying Lemma~\ref{lemma6_rav}, we have:
\begin{align}
\|\Delta\|_{\infty} \leq 2 \kappa_{\Gamma^{*}}\left(\|W\|_{\infty}+\lambda_{n}+ \|  (S^* - L^*)^{-1} -  (S^*)^{-1}   \|_{\infty} \right) \leq 2 \kappa_{\Gamma^{*}}\left(1+\frac{12}{\alpha}+C_2\right) \bar{\delta}_{f}\left(n, p^{\tau}\right).  \label{eq:64}
\end{align}

Now, for Lemma~\ref{lemma5_rav}, we see that the assumption $\|\Delta\|_{\infty} \leq \frac{1}{3 \kappa_{{S^*}^{-1}} d}$ holds by equation~\eqref{eq:63} and~\eqref{eq:64}. Therefore, we have:
$$
\begin{aligned}
\|R(\Delta)\|_{\infty} & \leq \frac{3}{2} d\|\Delta\|_{\infty}^{2} \kappa_{{S^{*}}^{-1}}^{3} \\
& \leq 6 \kappa_{{S^{*}}^{-1}}^{3} \kappa_{\Gamma^{*}}^{2} d\left(1+\frac{12}{\alpha}+C_2\right)^{2}\left[\bar{\delta}_{f}\left(n, p^{\tau}\right)\right]^{2} \\
&=\left\{6 \kappa_{{S^{*}}^{-1}}^{3} \kappa_{\Gamma^{*}}^{2} d\left(1+\frac{12}{\alpha}+C_2\right)^{2} \bar{\delta}_{f}\left(n, p^{\tau}\right)\right\} \frac{\alpha \lambda_{n}}{12} \\
& \leq \frac{\alpha \lambda_{n}}{12}.
\end{aligned}
$$

Overall, we have shown that the assumption~\eqref{eq:51} of Lemma~\ref{lemma4_rav} holds, allowing us to conclude that $\widetilde{\Theta}=\widehat{\Theta}$. The estimator $\widehat{\Theta}$ then satisfies the $\ell_{\infty}$-bound~\eqref{eq:64}  of $\widetilde{\Theta},$  and moreover, we have $\widehat{\Theta}_{S^{c}}=\widetilde{\Theta}_{S^{c}}=0$. By the $\ell_{\infty}$-bound and minimum signal strength condition,  we have sign consistency: $\operatorname{sign}(\widehat{\Theta}^{\lambda}_{ij})=\operatorname{sign}\left(S_{ij}^{*}\right), \forall S_{ij}^* \neq 0$. Since the above was conditioned on the event $\mathcal{A},$ these statements hold with probability $\mathbb{P}(\mathcal{A}) \geq 1-\frac{1}{p^{\tau-2}}$.
$\qed$

\section{Thresholded Graphical Lasso In the Presence of Latent Variables}\label{appen_tglasso_lv}
In this section, we study thresholded graphical Lasso in the presence of latent variables and consider conditions when thresholded graphical Lasso can yield a consistent estimate of the sparse concentration matrix $S$ in the latent variable graphical model.
 
\hide{ 
\begin{assumption}
\label{assumptionC1}
$X_{i} \text { be i.i.d. } \mathcal{N}\left(\mathbf{0}, \Sigma^* \right)$ where $\Sigma^* = (S^* - L^*)^{-1}$.
\end{assumption}
\begin{assumption}
\label{assumptionC2}
 $\varphi_{\min }\left( {S^*}^{-1} \right) \geq \underline{k}>0$, or equivalently $\varphi_{\max }\left(S^*\right) \leq 1 / \underline{k}$.
\end{assumption}
\begin{assumption}
\label{assumptionC3}
$\varphi_{\max }\left({S^*}^{-1}\right) \leq \bar{k}$.
\end{assumption}
}

\assumptionCone*
\assumptionCtwo*
\assumptionCthree*

\hide{
\begin{lemma}
\label{lemma2}
Let Assumptions~\ref{assumptionB4}-\ref{assumptionC3} be satisfied. If $\lambda \asymp \sqrt{\frac{\log p}{n}}$, there exists a  $\hat c_2$ such that the graphical Lasso estimate $\widehat{\bTheta}^{\lambda}$ satisfies:
$$
\left\| {\widehat{\bTheta}}^{\lambda}-S^{*}\right\|_{F} \leqslant \hat c_{2}  \sqrt{\frac{(p+s) \log p}{n}} \; ,
$$
with probability at least $1 - b_1 \exp(-b_2 n \lambda^2)$ where $b_1$ and $b_2$ depend on $\bar{k}$ only.
\end{lemma}
}
\lemmatwo*

Similarly, as mentioned in Lemma~\ref{lemma1},  the worst part of the rate, $\sqrt{p \log p /n}$, comes from estimating the diagonal. \hide{It can be shown that
$$
\mathbb{P}\left(  \left\|\widehat{\Theta}_{\text{off}}^{\lambda}-S^{*}_{\text{off}}\right\|_{F} \leq  c_{2}  \sqrt{\frac{s \log p}{n}}  \right) \to 1.
$$
We have:
\begin{proposition}
\label{proposition_rothman_off_diag_lv}
Let Assumptions~\ref{assumptionB4}-\ref{assumptionC3} be satisfied. If $\lambda \asymp \sqrt{\frac{\log p}{n}}$, there exists a  $c_2$ such that the graphical Lasso estimate $\widehat{\bTheta}^{\lambda}$ satisfies:
$$
\mathbb{P}\left(  \left\|\widehat{\Theta}_{\text{off}}^{\lambda}-S^{*}_{\text{off}}\right\|_{F} \leq  c_{2}  \sqrt{\frac{s \log p}{n}}  \right) \to 1.
$$
\end{proposition}
}
\propositionrothmanoffdiaglv*

Still, we further assume minimum signal strength condition to establish graphical model selection consistency.
\hide{
\begin{assumption}
\label{assumptionC4}
Define the minimum signal strength:
\begin{equation*}
\theta_{\min }:=\min _{(i, j) \in E\left(S^{*}\right)}\left|S_{i j}^{*}\right| > c_1 \sqrt{\frac{s \log p}{n}}.
\end{equation*}
\end{assumption}
}

\assumptionCfour*

\hide{
\begin{theorem}
\label{theorem3}
Let Assumptions~\ref{assumptionB4}-\ref{assumptionC4}  be satisfied. We assume furthermore that $c_1> 2c_2$, where $c_2$ is defined in 
Proposition~\ref{proposition_rothman_off_diag_lv}
Then the thresholded graphical Lasso estimate $\widetilde{\bTheta}^{\lambda,\tau}$ with threshold level $\tau = c_2 \sqrt{\frac{s \log p}{n}}$ satisfies:
\begin{align*}
    &\mathbb{P}\left(\operatorname{sign}(\widetilde{\bTheta}_{ij}^{\lambda,\tau})=\operatorname{sign}\left(S_{ij}^{*}\right), \forall \widetilde{\bTheta}^{\lambda,\tau}_{ij} \in \widetilde{\bTheta}^{\lambda,\tau}\right)  \geq 1  - b_1 \exp(-b_2 n \lambda^2) \to 1.
\end{align*}
\end{theorem}
}
\theoremthree*

We prove Lemma~\ref{lemma2} in the following; Theorem~\ref{theorem3} can be proved in the similar way as Theorem~\ref{theorem1}.

\noindent \underline{\textbf{Proof of Lemma~\ref{lemma2}:} }
We follow closely the proof approach used by \citet{rothman2008sparse}; we consider the effect of latent variables in our proof. Let
\begin{align}
Q(\bTheta)=& \operatorname{tr}(\bTheta \hat{\Sigma})-\log |\bTheta|+\lambda\left|\bTheta^{-}\right|_{1}-\operatorname{tr}\left(S^* \hat{\Sigma}\right)+\log \left|S^* \right|-\lambda\left| {S^*}^{-}\right|_{1} \nonumber\\
=& \operatorname{tr}\left[\left(\bTheta- S^*\right)\left(\hat{\Sigma}-(S^*)^{-1}  \right)\right]-\left(\log |\bTheta|-\log \left|S^*\right|\right) \nonumber\\
&+\operatorname{tr}\left[\left(\bTheta-S^*\right) (S^*)^{-1} \right]+\lambda\left(\left|\bTheta^{-}\right|_{1}-\left|{S^*}^{-}\right|_{1}\right). \label{eq:tglasso-1}
\end{align}

Our estimate $\widehat{\bTheta}$ minimizes $Q(\bTheta)$, or equivalently $\widehat{\Delta}=\widehat{\bTheta}-S^*$ minimizes $G(\Delta) \equiv$ $Q\left(S^*+\Delta\right)$. 
Consider the set
$$
\Theta_{n}(M)=\left\{\Delta: \Delta=\Delta^{T},\|\Delta\|_{F}=M r_{n}\right\},
$$
where
$$
r_{n}=\sqrt{\frac{(p+s) \log p}{n}} \rightarrow 0.
$$
Note that $G(\Delta)=Q\left(S^* +\Delta\right)$ is a convex function, and
$$
G(\widehat{\Delta}) \leq G(0)=0.
$$
Then, if we can show that
$$
\inf \left\{G(\Delta): \Delta \in \Theta_{n}(M)\right\}>0,
$$
the minimizer $\widehat{\Delta}$ must be inside the sphere defined by $\Theta_{n}(M)$, and hence
$$
\|\widehat{\Delta}\|_{F} \leq M r_{n}.
$$
For the logarithm term in \eqref{eq:tglasso-1}, we use Taylor expansion of $f(t)=\log |\bTheta+t \Delta|$ and integral form of the remainder:
$$
\log \left|S^*+\Delta\right|-\log \left| S^* \right|=\operatorname{tr}\left( (S^*)^{-1} \Delta\right)-\tilde{\Delta}^{T}\left[\int_{0}^{1}(1-v)\left( S^* +v \Delta\right)^{-1} \otimes\left(S^* +v \Delta\right)^{-1} d v\right] \tilde{\Delta},
$$
where $\otimes$ is the Kronecker product 
and $\tilde{\Delta}$ is $\Delta$ vectorized to match the dimensions of the Kronecker product.

Therefore, we may write \eqref{eq:tglasso-1} as,
$$
\begin{aligned}
G(\Delta)=& \operatorname{tr}\left(\Delta\left(\hat{\Sigma} - (S^*)^{-1} \right)\right)+\tilde{\Delta}^{T}\left[\int_{0}^{1}(1-v)\left(S^* +v \Delta\right)^{-1} \otimes\left( S^* +v \Delta\right)^{-1} d v\right] \tilde{\Delta} \\
&+\lambda\left(\left|{S^*}^{-}+\Delta^{-}\right|_{1}-\left|{S^*}^{-}\right|_{1}\right).
\end{aligned}
$$
By triangular inequality,
$$
\left|\operatorname{tr}\left(\Delta\left(\hat{\Sigma}-(S^*)^{-1} \right)\right)\right| \leq\left|\sum_{i \neq j}\left(\hat{\sigma}_{i j}-\sigma_{s i j}\right) \Delta_{i j}\right|+\left|\sum_{i}\left(\hat{\sigma}_{i i}-\sigma_{s i i}\right) \Delta_{i i}\right|=\mathrm{I}+\mathrm{II},
$$
where we denote $(S^*)^{-1}_{ij} = \sigma_{s i j}$. Further, we denote $\hat{\Sigma}_{ij} = \hat \sigma_{ij}$, ${\Sigma}^*_{ij} =  \sigma_{0ij}$ and $\left[(S^* - L^*)^{-1} -  (S^*)^{-1}\right]_{ij} = \eta_{ij}$.

Note that
\begin{align*}
    \hat{\Sigma}-(S^*)^{-1} &= \hat{\Sigma}- \Sigma^* + \Sigma^* -  (S^*)^{-1} \\
    &= \hat{\Sigma}- \Sigma^* + (S^* - L^*)^{-1} -  (S^*)^{-1},
\end{align*}
where $(S^* - L^*)^{-1} -  (S^*)^{-1}$ is the same term we establish in Assumption~\ref{assumptionB4}.

To bound term I, note that the union sum inequality and Lemma 1 of \citet{rothman2008sparse} (also Lemma 3 of \citet{bickel2008regularized}) imply that, with probability tending to 1,
$$
\max _{i \neq j}\left|\hat{\sigma}_{i j}-\sigma_{0 i j}\right| \leq C_{1} \sqrt{\frac{\log p}{n}}.
$$
Hence if we assume $\eta_{ij} = \|  (S^* - L^*)^{-1} -  (S^*)^{-1}   \|_{\infty, \text{off}} \leq \tilde C_1  \sqrt{\frac{\log p}{n}}$, term I is bounded by
$$
\mathrm{I} \leq \max_{i \neq j} \left|\hat{\sigma}_{i j}-\sigma_{s i j}\right| \cdot \left|\Delta^{-}\right|_{1} \leq (C_1+ \tilde C_1) \sqrt{\frac{\log p}{n}}\left|\Delta^{-}\right|_{1}.
$$

The second bound comes from the Cauchy-Schwartz inequality and Lemma 1 of \citet{rothman2008sparse}. We assume  $\eta_{ii} = \|  (S^* - L^*)^{-1} -  (S^*)^{-1}   \|_{\infty, \text{diag}} \leq \tilde C_2  \sqrt{\frac{\log p}{n}}$. We have
\begin{align*}
\mathrm{II} & \leq\left[\sum_{i=1}^{p}\left(\hat{\sigma}_{i i}-\sigma_{s i i}\right)^{2}\right]^{1 / 2}\left\|\Delta^{+}\right\|_{F} \leq \sqrt{p} \max _{1 \leq i \leq p}\left|\hat{\sigma}_{i i}-\sigma_{s i i}\right|\left\|\Delta^{+}\right\|_{F} \\
& \leq (C_2 + \tilde C_2) \sqrt{\frac{p \log p}{n}}\left\|\Delta^{+}\right\|_{F} \leq (C_2 + \tilde C_2) \sqrt{\frac{(p+s) \log p}{n}}\left\|\Delta^{+}\right\|_{F},
\end{align*}
also with probability tending to 1.

The rest of the proof follows as \citet{rothman2008sparse}  paper.
$\qed$

\section{Thresholded Neighborhood Selection and CLIME In the Presence of Latent Variables}\label{appen_tns_lv}
 
In this section, we demonstrate the graph selection consistency of thresholded neighborhood selection and CLIME in the presence of latent variables.

We can easily extend Proposition~\ref{propositionNB2} by noting the true covariance matrix now becomes $\Psi^* = (S^*-L^*)^{-1}$. Hence, we can yield graphical model selection consistency of neighborhood selection  by letting Assumption~\ref{assumptionNB2} hold true for $\Psi^* = (S^*-L^*)^{-1}$.

 
To investigate thresholded CLIME in the presence of latent variables, \citet{ren2012} proposed a procedure to obtain an algebraically consistent estimate of the latent variable graphical model based on (thresholded) CLIME estimator. For completeness, we here restate their theory for thresholded CLIME in the presence of latent variables. In particular, they required that $\| L^* \|_{\infty} \leq c \sqrt{\frac{\log p}{n}}$, 
a similar assumption we have in Assumption~\ref{assumptionB4}.

\begin{proposition}[\citealt{ren2012}]
\label{proposition2}
Suppose that $S^{*} \in \mathcal{U}\left(s_{0}(p), M_{p}\right)$,
$$\sqrt{(\log p) / n}=o(1) \quad \text{ and } \quad\left\|L^{*}\right\|_{\infty} \leq M_{p} \tau_{n}.$$
With probability greater than $1-C_{s} p^{-6}$ for some constant $C_{s}$ depending on $M$ only, we have
$$
\left\|\widehat{S}-S^{*}\right\|_{\infty} \leq 9 M_{p} \tau_{n}.
$$
Hence, if the minimum magnitude of nonzero entries $\theta>18 M_{p} \tau_{n}$, we obtain the sign consistency $\operatorname{sign}(\widetilde{S})=\operatorname{sign}\left(S^{*}\right)$. In particular, if $M_{p}$ is in the constant level, then to consistently recover the support of $S^{*}$, we only need that $\theta \asymp \sqrt{(\log p) / n}$.
\end{proposition}

\section{Additional Empirical Studies to Select Regularization $\lambda$ and Threshold Level $\tau$}\label{appen_heatmap}

In this section, we show how different combinations of  regularization $\lambda$ and threshold level $\tau$ affect edge recovery and propose ways to select $\lambda$ and $\tau$ when the oracle number of edges is known. When the oracle number of edges is unknown, we propose to adopt the approaches discussed in Section 3.4.1 to choose the optimal combination of regularization $\lambda$ and threshold $\tau$.

\begin{figure}[ht]
\vskip 0.2in
\begin{center}
\centerline{\includegraphics[scale = 0.9]{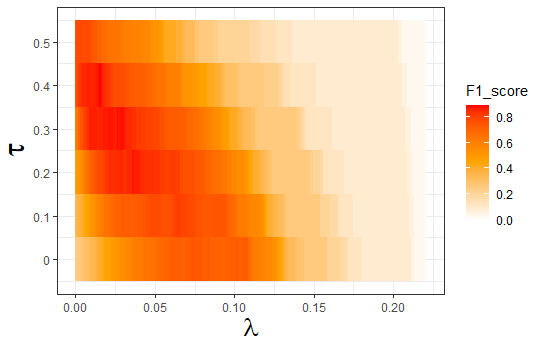}}
\caption{Heatmap of F1-score for combinations of regularization $\lambda$ and level of threshold $\tau$. We show how edge recovery varies with respect to different combinations of $\lambda$ and $\tau$. We measure edge recovery accuracy in terms of F1-score. Here we scale the level of threshold $\tau$ by  $\|\hat \Theta\|_{\infty,\text{off}}$.}
\label{heatmap}
\end{center}
\vskip -0.2in
\end{figure}

We find that edge recovery is robust to the choice of $\lambda$ when proper level of threshold $\tau$ is then chosen to give sparse graph, as long as $\lambda$ is sufficiently small (and therefore produces dense solutions), as shown in Figure~\ref{heatmap}. Hence, we propose to fit regularized graphical model with an initial small $\lambda_0 \propto \sqrt{\frac{\log p}{n}}$ and then choose the level of threshold which gives oracle number of edges. Also, Figure~\ref{heatmap} suggests that thresholding a dense solution is better than just using a sparse solution (large $\lambda$ with $\tau = 0$).

\end{appendix}

\bibliographystyle{abbrvnat}
\bibliography{main.bib}

\end{document}